%% file: main.tex
\documentclass[acmsmall,review=false]{acmart}

\AtBeginDocument{%
  \providecommand\BibTeX{{%
    \normalfont B\kern-0.5em{\scshape i\kern-0.25em b}\kern-0.8em\TeX}}}

\setcopyright{acmcopyright}
\copyrightyear{2020}
\acmYear{2019}
\acmDOI{10.1145/1122445.1122456}

\acmJournal{JACM}
\acmVolume{01}
\acmNumber{01}
\acmArticle{01}
\acmMonth{01}

\usepackage{graphicx}
\usepackage{float}
\usepackage{enumitem}
\usepackage{tabu}
\usepackage{url}
\usepackage{bm}
\usepackage{epstopdf}
\usepackage{subfigure}
\usepackage{multirow}
\usepackage[linesnumbered,algoruled,boxed,lined]{algorithm2e}
\usepackage{amsmath,amssymb,amsfonts}
\usepackage{algorithmic}
\usepackage{textcomp}
\usepackage{xcolor}
\usepackage{makecell}
\DeclareMathOperator*{\argmax}{arg\,max}

\newtheorem{mydefinition}{Definition}
\begin{document}

\title{Developing Multi-Task Recommendations with Long-Term Rewards via Policy Distilled Reinforcement Learning}

\author{Xi Liu}
\authornote{Both authors contributed equally to this research.}
\email{xiliu.tamu@gmail.com}
\affiliation{%
  \institution{Texas A\&M University}
  \city{College Station}
  \state{Texas}
  \postcode{77840}
}

\author{Li Li}
\authornotemark[1]
\email{li.li1@samsung.com}
\affiliation{%
  \institution{Samsung Research America}
  \city{Mountain View}
  \state{California}
  \postcode{94043}
}

\author{Ping-Chun Hsieh}
\email{pinghsieh@cs.nctu.edu.tw}
\affiliation{%
  \institution{National Chiao Tung University}
  \city{Hsinchu}
  \state{Taiwan}
}

\author{Muhe Xie}
\affiliation{%
  \institution{Samsung Research America}
  \city{Mountain View}
  \state{California}
  \postcode{94043}}
\email{muhexie@gmail.com}

\author{Yong Ge}
\affiliation{%
  \institution{University of Arizona}
  \city{Tucson}
  \state{Arizona}
  \postcode{85721}}
\email{yongge@email.arizona.edu}

\author{Rui Chen}
\affiliation{%
  \institution{Samsung Research America}
  \city{Mountain View}
  \state{California}
  \postcode{94043}}
\email{rui.chen1@samsung.com}

\renewcommand{\shortauthors}{Liu and Li, et al.}

\input 0_abstract


\maketitle

\input 1_introduction

\input 2_related_work

\input 3_problem_statement

\input 4_methods

\input 5_experiments

\input 6_conclusion

\input 7_appendix

\bibliographystyle{unsrtnat}
\bibliography{ref}

\end{document}

%% file: 0_abstract.tex
\begin{abstract}
  With the explosive growth of online products and content, recommendation techniques have been considered as an effective tool to overcome information overload, improve user experience, and boost business revenue. In recent years, we have observed a new desideratum of considering long-term rewards of multiple related recommendation tasks simultaneously. The consideration of long-term rewards is strongly tied to business revenue and growth. Learning multiple tasks simultaneously could generally improve the performance of individual task due to knowledge sharing in multi-task learning. While a few existing works have studied long-term rewards in recommendations, they mainly focus on a single recommendation task. In this paper, we propose {\it PoDiRe}: a \underline{po}licy \underline{di}stilled \underline{re}commender that can address long-term rewards of recommendations and simultaneously handle multiple recommendation tasks. This novel recommendation solution is based on a marriage of deep reinforcement learning and knowledge distillation techniques, which is able to establish knowledge sharing among different tasks and reduce the size of a learning model. The resulting model is expected to attain better performance and lower response latency for real-time recommendation services. In collaboration with Samsung Game Launcher, one of the world's largest commercial mobile game platforms, we conduct a comprehensive experimental study on large-scale real data with hundreds of millions of events and show that our solution outperforms many state-of-the-art methods in terms of several standard evaluation metrics.
\end{abstract}

%% file: 1_introduction.tex
\section{Introduction}\label{sec:introduction}
    
With the explosive growth of online information, users are often greeted with more than countless choices of products and content. For example, as of the first quarter of 2019, there were over 3.9 million active apps in the Apple App Store and Google Play~\cite{StatistaReport}; there are billions of items on e-commerce websites like Amazon~\cite{AmazonReport} and billions of videos on video-sharing websites like YouTube~\cite{YoutubeReport}. Recommender systems have consequently become an effective and indispensable tool to overcome information overload, boost stakeholders' revenue and, improve user experience~\cite{ChangBiaoCIKM2014}. Nowadays, recommendation techniques have been widely studied and deployed in a wide range of application domains. 
    
Traditional recommendation techniques~\cite{baltrunas2011matrix, rendle2012factorization,verbert2012context,bobadilla2013recommender,wang2015collaborative,cheng2016wide,xiong2017end,he2017collaborative,tang2018personalized,ZhaoKDD20161,chen2019joint,xue2019deep} usually focus on estimating and maximizing immediate (short-term) rewards of current recommendations (e.g., immediate clicks on recommended apps), and largely overlook \textit{long-term rewards} driven by current recommendations (e.g., clicks on and/or installations of future recommended apps). The long-term rewards of recommendations are essentially the positive impact of current recommendations on users' responses to future recommendations~\cite{chen2019top,wang2018supervised,chen2018}. Such long-term rewards are now considered as a top business desideratum because they are strongly tied to revenue and growth. To this end, a recently-developed branch of methods started to look into long-term rewards of recommendations~\cite{Yu2019,Lei2019,Ie2019,chen2019large,chen2019top,zhao2018,wang2018supervised,chen2018,zheng2018drn,oosterhuis2018,zhao2018drl}. These studies have demonstrated that modeling long-term rewards could greatly improve the overall performance of recommender systems, users' lifetime values, and long-term business revenue.

    \begin{figure}[t]
      \centering
      \includegraphics[width=0.7\linewidth]{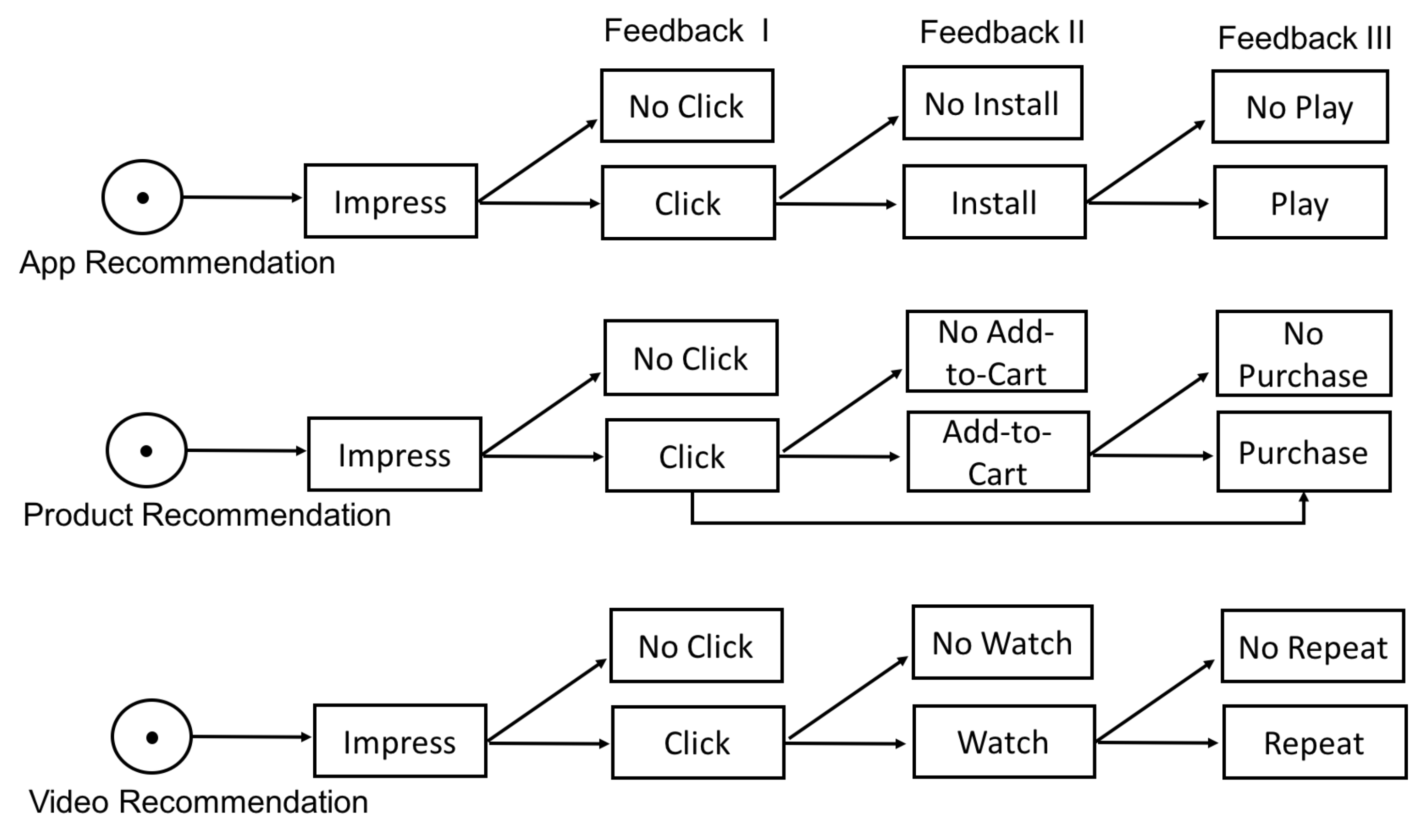}
      \caption{Multiple types of recommendation feedback in different applications}
      \label{fig:muptile_feedbacks}
    \vspace{-6mm} 
    \end{figure}
    
Most of these existing studies on long-term rewards of recommendations consider an individual recommendation task that aims to optimize a single type of user responses such as click. However, in many real-world business applications, we often simultaneously face \textit{multiple} recommendation tasks, each of which targets to optimize one type of user feedback. Fig.~\ref{fig:muptile_feedbacks} illustrates three examples of multiple-type user feedback to be optimized in different applications: click, install, and play for game app recommendations~\cite{liu2018semi,Liu2019}, click, add-to-cart, and purchase for online product recommendations~\cite{pei2019value}, and click, watch, and repeat for online video recommendations~\cite{covington2016deep}. The multiple types of response in Fig.~\ref{fig:muptile_feedbacks} are generated when users interact with recommended items and provide different levels of feedback. A single recommendation task could be formulated with respect to each type of feedback. For instance, with Feedback I (i.e., click) shown in Fig.~\ref{fig:muptile_feedbacks}, one single recommendation task is to recommend items (e.g., app, product or video) that are likely to be clicked to users. Although it is possible to solve each single recommendation task separately, it is much more promising to consider multiple recommendation tasks in a joint way for the following reasons. First, there is natural dependency among different types of feedback, as well as the corresponding recommendation objectives~\cite{ahmed2014scalable}. For instance, for game app recommendations shown in Fig.~\ref{fig:muptile_feedbacks}, a user may first click a recommended app,  then download the app to mobile terminals, and finally play the game. Effectively modeling the dependency among these types of feedback and objectives will promote knowledge sharing and promisingly improve the performance of all recommendation tasks. Second, a single type of feedback could be highly sparse and imbalanced~\cite{ma2018esm}, making the development of a single-task recommendation model highly challenging, though resampling techniques could be used before training~\cite{ma2018esm}. Learning common feature representations and jointly considering these recommendation tasks have been shown to be effective in relieving the data sparsity and imbalance issue~\cite{zhang2017survey}.  
    
In addressing the need for solving multiple recommendation tasks together and maximizing long-term rewards, we propose {\it PoDiRe}, a \underline{po}licy \underline{di}stilled \underline{re}commender that is able to efficiently and simultaneously handle multiple long-term-reward maximization tasks, each of which corresponds to one type of user feedback. To take into account the long-term impacts of current recommendation to subsequent rewards, {\it PoDiRe} models the interaction between the recommender and a user by a Markov decision process (MDP), where a recommendation may incur state transition of the user. Consequently, the user's subsequent responses to later recommendations will be based on the transited state. {\it PoDiRe} adapts deep neural network (DNN) to automatically learn state representation as well as the optimal (expert) policy for each task. To facilitate knowledge sharing, improve the performance over multiple tasks and attain a compact model, {\it PoDiRe} combines multiple expert policies learned for different tasks into a single multi-task policy that can outperform the separate experts. Such a process is known as policy distillation~\cite{andrei2016}. In {\it PoDiRe}, the DNN trained to capture the expert policy for a single task is called teacher network, and the DNN that represents the multi-task policy is named student network.

Each teacher network is encoded by a Double Deep Q-Network (DDQN), which is a more stable and robust variant of deep Q-learning method~\cite{hasselt2016deep}. In most previous studies~\cite{zhao2018,oosterhuis2018,zhao2018drl}, the representation of the state for the teacher network is usually obtained from the $T$  most recent recommendations a user has interacted with, where $T$ is a hyper-parameter determined by cross-validation. In the real practice of Samsung Game Launcher recommender, we noticed that the state represented in these ways might change abruptly in two consecutive recommendations, making the training of the model unstable. To resolve this issue, we introduce into the state a relevant and relatively static part that summarizes information from all historical interactions, app usage, as well as user profiles. This first part is referred to as ``long-term interest'' of the user, and the second part, similar to previous methods, is referred to as ``short-term interest''. Finally, to take advantage of the relevance between contexts (e.g., time and location) and user feedback, contexts of one recommendation are also included in the state as the third part.
    
The student network is encoded by multi-layer feedforward neural networks with task-sharing and task-specific layers. Through the task-sharing layers, common feature representations are learned among different tasks to resolve the data sparsity and imbalance. Thanks to the extra data from distillation, the size of the student network is compressed. By optimizing a well-designed multi-task loss, the student network encourages knowledge sharing among different tasks. As a result, its performance in the testing phase outperforms all teacher networks over each individual task. In the testing (or serving) phase, we can actually discard the teacher networks and only use the student network to simultaneously generate recommendations for multiple tasks, and thus the latency of responding to user requests can be significantly reduced.

Our overall research contributions are summarized in the following.
\begin{enumerate}
    \item To the best of our knowledge, this is the first study to develop a solution to simultaneously solve multiple recommendation tasks with the goal of maximizing long-term rewards. Our solution meets important business needs in many real-world applications. 
    
    \item We propose a novel multi-task recommendation framework based on policy distillation, which includes multiple teacher networks and a student network. This is the first solution to apply policy distillation from multiple recommendation tasks. It allows us to obtain a model with a smaller size and lower response latency, making it more appealing for real-world deployments.
    
    \item We also design a different state representation method than previous studies to make the training of our model more stable. The state of a user in a recommendation is represented by three parts: the user's short-term interest, the user's long-term interest as well as the rich context information of recommendation.
    
    \item We conduct a comprehensive experimental evaluation over the Samsung Game Launcher platform, which is one of the largest commercial mobile game platforms. The experiments consists of hundreds of millions of real-world log events. The experimental results demonstrate that our model outperforms many state-of-the-art methods based on multiple evaluation metrics.
\end{enumerate}

The rest of this paper unfolds in the following order. We review related works in Section \ref{sec:related}. Section \ref{sec:problem} introduces the problem statement. To address this problem, an innovative method (i.e., {\it PoDiRe}) is provided in Section \ref{sec:methods}. Comprehensive evaluations and main findings using a real-world dataset are provided in Section \ref{section:experiment}. Finally, in Section \ref{section:conclusion} we draw our conclusion.

%% file: 2_related_work.tex
\section{Related Work}\label{sec:related}

Advanced machine learning methods have been developed for solving various recommendation problems~\cite{LiHuayuICDM2015,LeWuAAAI2016,baltrunas2011matrix,rendle2012factorization,FuSIAMSDD20141,lian2018geomf++,loni2019top,arampatzis2017suggesting}. These methods are often based on supervised learning (e.g., classification, prediction, etc.), where user feedback is viewed as labels to be classified or predicted. Their models could be further grouped into two categories: linear and non-linear. Representative linear methods include logistic regression (LR)~\cite{tang2016empirical}, matrix factorization (MF)~\cite{baltrunas2011matrix} and factorization machines (FMs)~\cite{rendle2010factorization,rendle2012factorization}. MF represents a user or an item by a vector of latent features and models a user-item interaction by the inner product of their latent vectors. FMs embed features into a latent space and model user-item interactions by summing up the inner products of embedding vectors between all pairs of features. The inner products, as well as the sum, simply combine the multiplication of latent features linearly. Such linear operations might be insufficient to capture the inherent non-linear and complex structure of real-world data. Therefore, more and more recent efforts have been invested in modeling user-item interactions by DNNs, which form the category of non-linear models. Sample works of this category include a mobile recommendation system based on gradient boosting decision tree (GBDT)~\cite{wang2016mobile}, the Wide\&Deep (W\&D) model for app recommendation in Google Play~\cite{cheng2016wide}, the non-linear extensions of MF and FM~\cite{he2017factorization,he2017collaborative}, the recommender modeled by recurrent neural networks (RNNs)~\cite{donkers2017sequential}, etc. A comprehensive review of those studies is available in~\cite{zhang2019deep}. However, the majority of these methods are designed to estimate and maximize immediate rewards of recommendations, neglecting long-term effects of current recommendation to subsequent rewards and thus unable to strategically maximizes long-term rewards.

To address the long-term rewards, reinforcement learning (RL) has been applied to different recommendation tasks, including video recommendation~\cite{Ie2019,chen2019top,chen2019large,Lei2019}, e-commerce recommendation~\cite{zhao2018,zhao2018drl,chen2018,oosterhuis2018,theocharous2015}, news recommendation~\cite{zheng2018drn}, and treatment recommendation~\cite{wang2018supervised}. Compared to conventional techniques, RL models consider that the rewards of recommendation are state-dependent, the current recommendation incurs state transition and the next recommendation will be made on the transited state. In this way, RL models aim to learn an action policy for an agent (e.g., the recommender) to maximize the expected long-term rewards in a sequence of interactions between the agent and the environment (e.g., the user)~\cite{sutton2018reinforcement}. Note that different from the supervised learning based methods, the expected long-term rewards are not initially available like ``labels'' and thus has to be first estimated in training the RL models. This brings more complexity in training. Compared with previous state-of-the-art methods, these RL-based recommenders yield better performance in terms of several evaluation metrics. However, these recommenders are designed to perform a single recommendation task that optimizes a single type of user feedback and thus cannot jointly handle multiple recommendation tasks, each of which optimizes a different type of user feedback. This real-world driven problem is the focus of this paper. 

Multi-task learning (MTL) is a learning paradigm in machine learning that aims to leverage useful information shared in multiple related tasks to help improve the performance of all the tasks~\cite{zhang2017survey}. Recently a few studies have applied MTL to ``mixed'' tasks that include recommendation and other non-recommendation tasks. These studies can be summarized into two categories. The first category includes studies that address the main task and an auxiliary task~\cite{bansal2016ask,li2017neural}. The main task and the auxiliary task are trained jointly to improve the performance of the main task, and the performance of the auxiliary task is usually not the focus. For example, Bansal \textit{et al.}~\cite{bansal2016ask} utilize item genre prediction as an auxiliary task to improve the performance of the main recommendation task. The second category contains works that have no priority among different tasks and aim to optimize the performance of all tasks~\cite{lu2018like,jing2017neural}. For instance, Jing \textit{et al.}~\cite{jing2017neural} simultaneously solve the user returning time prediction task and the recommendation task. None of these prior works address multiple recommendation tasks simultaneously, let alone considering long-term rewards of recommendations.

%% file: 3_problem_statement.tex
\section{Problem Statement}\label{sec:problem}
The interaction between a recommender and a user in a single recommendation task can be considered as a sequential decision process, i.e., the recommender deciding a sequence of recommendations to the user. We model the sequential process by a Markov decision process (MDP), which could enable the recommender to maximize long-term rewards. In the MDP, a user is considered as the ``environment'' represented by a ``state'', a recommender as the ``agent'', a recommended item as the ``action'' from the agent to the environment, and a response as the immediate ``reward'' of the action returned from the environment. The fundamental assumption in the MDP is that an action may incur state transition of the environment, and consequently, the agent will make a decision at the next time step based on the transited state. In this way, we impose some consideration of future rewards in current decision making. For example, if an action makes the environment transit to a state without any future rewards (e.g., all future feedback is negative), it will be disregarded even if the immediate reward is positive.

\begin{figure}[t]
  \centering
  \includegraphics[width=0.5\linewidth]{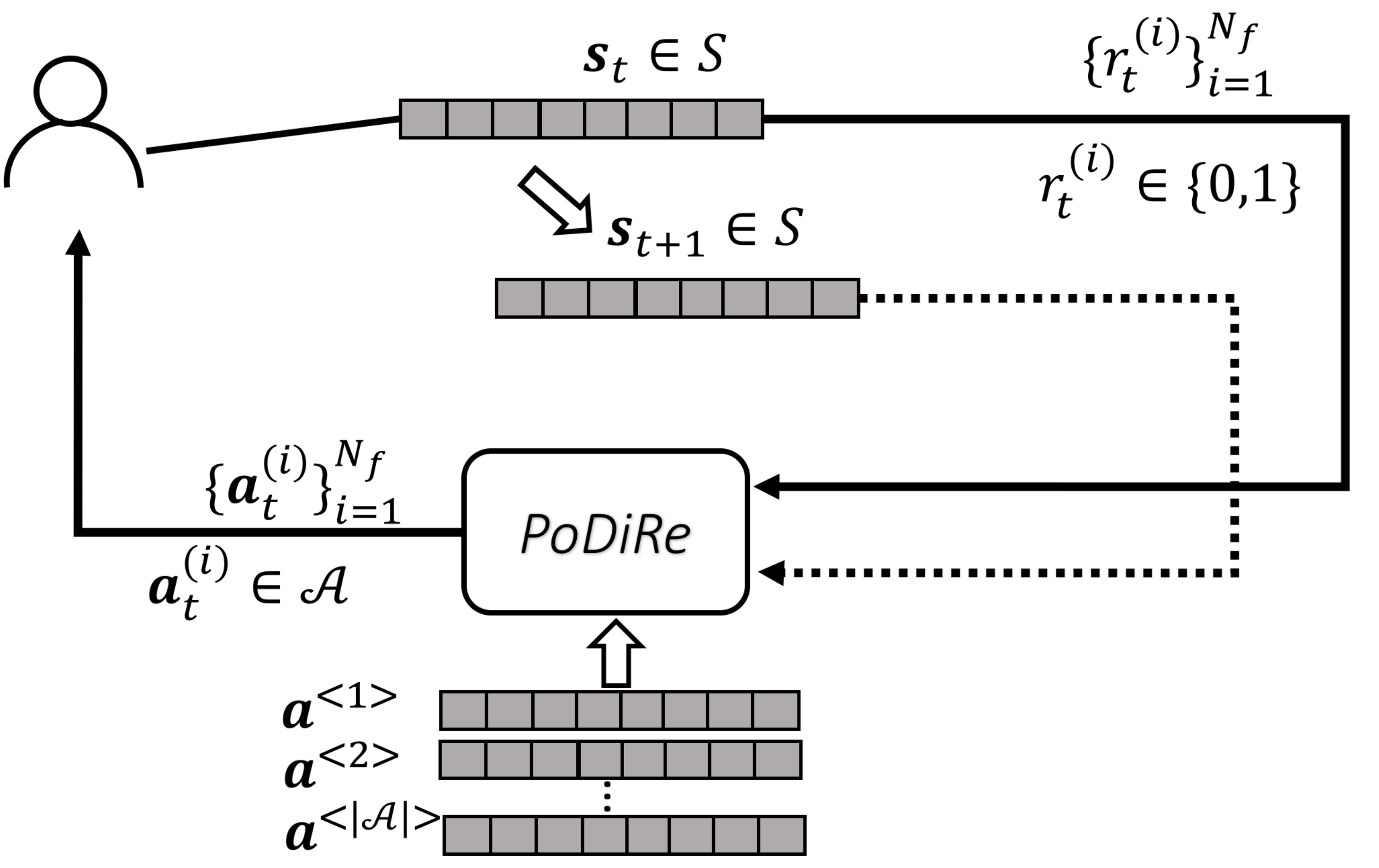}
  \caption{An example of user-{\it PoDiRe} interactions}
  \label{fig:overview}
  \vspace{-4mm} 
\end{figure}

A basic MDP suffices to model a recommender for a single task and can be solved by basic Reinforcement Learning (RL). But it will fail when more than one task needs to be handled. This is largely because such an RL-based recommender is designed to make a decision to optimize a single type of long-term rewards. In contrast, in the multi-task setting, multiple decisions need to be made to simultaneously optimize different types of long-term rewards. To obtain a multi-task recommender, we extend the recommender modeled by the basic RL to {\it PoDiRe}, in which more than one action is taken on any state and each of the actions optimizes a single type of long-term rewards. In recommendations, one type of rewards usually corresponds to one kind of user feedback, such as clicks or installations to app recommendations. We illustrate the interactions between a user and the {\it PoDiRe} in Fig.~\ref{fig:overview}. At the $t$-th recommendation, where $t=1,2,\cdots$ is the arrival order of recommendation requests, the preference of a user before the $t$-th recommendation is encoded by state $\bm{s}_{t} \in \mathcal{S}$, where $\bm{s}_{t}$ is a vector and $\mathcal{S}$ is the state space. Given $N_{f}$ recommendation tasks, to optimize the $i$-th one, {\it PoDiRe} recommends item $\bm{a}^{(i)}_{t} \in \mathcal{A}$ to the user and observes $r^{(i)}_{t} \in \{0,1\}$, the reward of type $i$ feedback, from the user (e.g., $r^{(i)}_{t}=1$ for click and $0$ for no click), where $\bm{a}^{(i)}_{t}$ is a vector representing the action taken in $i$-th task at $t$-th recommendation and $\mathcal{A}$ is the set containing of possible actions. The $k$-th action corresponds to an item and is represented by a vector $\bm{a}^{<k>}$, where $k \in \{1, 2,...,|\mathcal{A}|\}$ denotes the action index among all actions contained in $\mathcal{A}$. To avoid confusion, we use superscript $(i)$ to refer to the $i$-th task, and the subscript $t$ for the $t$-th recommendation. When there is no subscript or superscript, $\bm{a}$ and $\bm{s}$ refer to an arbitrary action or state. Based on the item recommended and the multiple types of feedback obtained at $t$-th recommendation, the state of the user is considered to transit to $\bm{s}_{t+1}$ at the next recommendation upon the $(t+1)$-th request. Since the received feedback (the value of $r^{(i)}_{t}$) is stochastic, this transition is usually not deterministic and is thus modeled by a probability distribution $\mathcal{P}^{(i)}$ whose probability is 
$P^{(i)}(\cdot | \cdot, \cdot):\mathcal{S}\times \mathcal{A} \times \mathcal{S} \to [0, 1]$, e.g., $P^{(i)}(\bm{s}_{t+1} | \bm{s}_{t}, \bm{a}_{t})$ is the probability that the user transits to state $\bm{s}_{t+1}$ after the agent takes action $\bm{a}_t$ at state $\bm{s}_{t}$ in the $i$-th task. The probability distribution $\mathcal{P}^{(i)}$ in practice is usually unknown but ``observable'', i.e., we can observe state transitions under different actions and have empirical estimation to the transition probability. Based on the above statement, we formulate the problem of multi-task recommendation with long-term rewards as follows.

\begin{mydefinition}\label{mydefinition} (\textit{Multi-Task Recommendation with Long-term Rewards}): Given the state space $\mathcal{S}$, the action set $\mathcal{A}$, the observable probability distribution for state transition $\mathcal{P}$, and the user (environment) that can provide immediate reward to actions, the goal is to find a unified multi-task recommendation policy, denoted by $\bm{\pi}_{S} = \{\pi_{S}^{(i)}(\cdot)\}_{i=1}^{N_f}$, such that $\pi_{S}^{(i)}(\cdot):\mathcal{S} \to \mathcal{A}$ can generate recommendation at any state in $\mathcal{S}$ and maximizes the long-term rewards in $i$-th recommendation tasks.
\end{mydefinition} 

The multi-task policy is learned from a set of sequences of interactions (recommendations and multiple types of feedback) $\big\{(\bm{s}_{t}, \{\bm{a}^{(i)}_{t}\}_{i=1}^{N_f}, \bm{s}_{t+1}, \{r^{(i)}_{t}\}_{i=1}^{N_{f}})\big\}_{t=1}^{N}$, that can be collected as the user-recommender interactions go by. Note that the subscript $S$ in $\bm{\pi}_{S}$ indicates that the policy is the student policy. The details about the student policy and teacher policies will be elaborated in Section~\ref{sec:methods}. A seemingly straightforward way to handle multiple types of feedback is to first combine rewards of different types of feedback into a weighted sum and then maximize the combined long-term reward.  It is worth noting our problem formulated in Definition~\ref{mydefinition} is different than this straightforward method because it essentially considers only a single recommendation task (i.e., generating recommendations by optimizing a single objective) even though the objective is obtained by combining multiple types of feedback. In contrast, Definition~\ref{mydefinition} considers the desideratum of jointly handling more than one recommendation task, i.e., simultaneously generating multiple recommended items towards the optimization of multiple objectives. Moreover, the straightforward method fails to take advantage of the shared knowledge and relatedness between multiple recommendation tasks, let alone improving the recommender's performance on multiple tasks by solving them jointly.

%% file: 4_methods.tex
\section{Research Method}\label{sec:methods}

\input 4_1_overview_of_methods

\input 4_2_teacher_model

\input 4_3_student_model

\input 4_4_state_action_representation

%% file: 4_1_overview_of_methods.tex
\subsection{An Overview of Proposed Method}

We utilize RL to optimize the long-term rewards of our recommendations. At a given state, RL-based recommenders recommend the item with the largest expected long-term rewards at this sate, which is usually referred to as the largest ``state-action value''. Integrating the multi-task learning (MTL) capability into such RL-based recommenders is more challenging than into conventional recommenders that are based on supervised learning (SL) techniques such as classification and regression models. First, different from SL, where the target to be learned (i.e., the ground-truth label) is given and fixed,  the state-action values in RL are not given or fixed and have to be estimated during training. Unfortunately, the learning of the state-action values can be quite unstable due to the exploration-and-error process~\cite{mnih2015human}. The instability will be amplified when more than one task (i.e., multiple tasks) is involved jointly. For instance, the gradients of one task may interfere with the learning of another task, or in the extreme, dominate the others, resulting in a negative impact on the performance of each other~\cite{andrei2016,parisotto2016actor}. Second, the state-action values of an RL-based recommender are normally real-valued and unbounded, and their scales can be different among tasks. MTL usually relies on a number of similar tasks as a shared source of inductive bias to improve generalization performance~\cite{caruana1997multitask,zhou2011multi,zhou2011clustered}. The varying and unbounded scales of individual tasks undermine the common statistical basis that MTL desires, making it more challenging to apply most existing MTL frameworks that expect similar scales in multiple tasks. Due to these challenges, most existing multi-task recommenders are built upon SL methods, rather than RL ones.

\begin{figure}[t]
  \centering
  \includegraphics[width=0.7\linewidth]{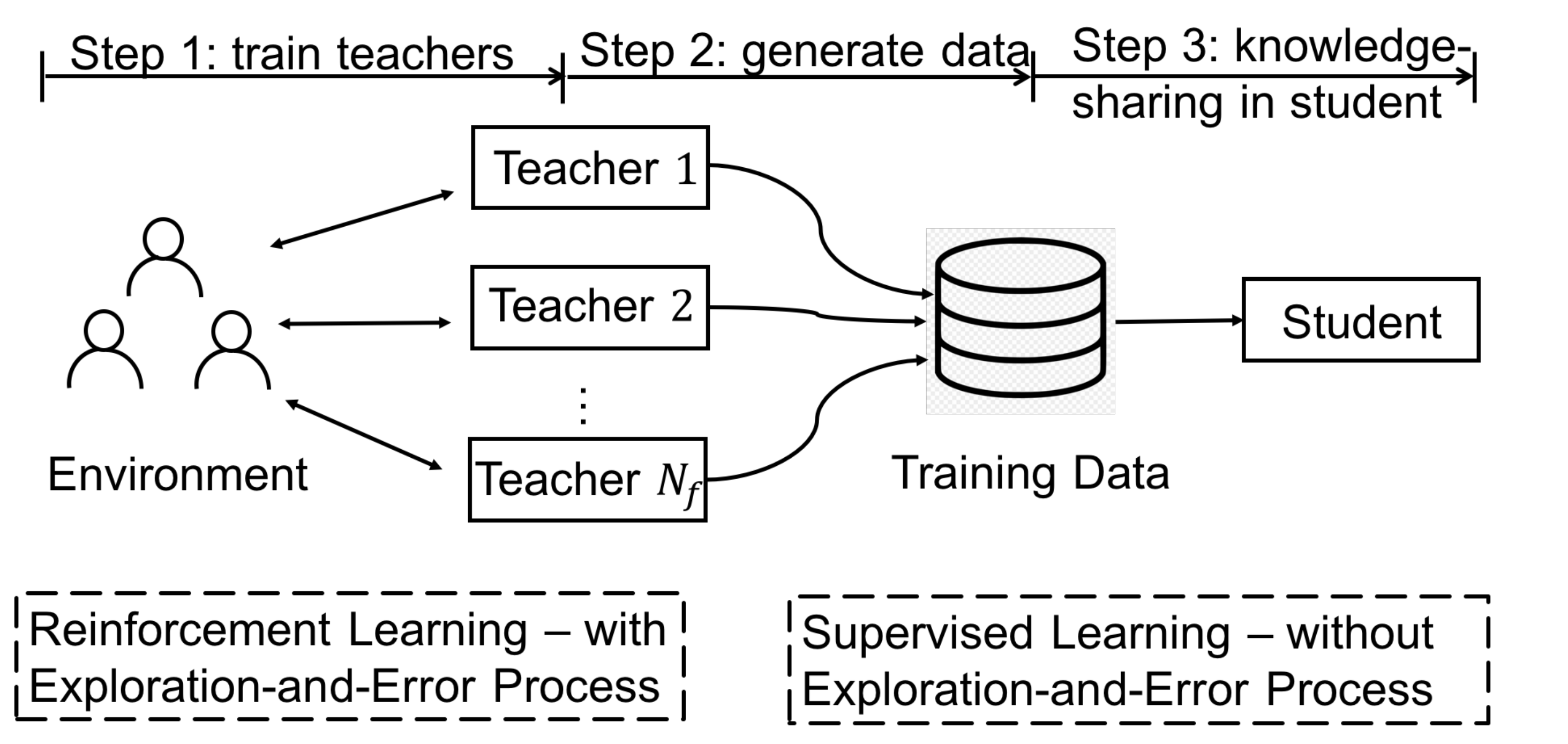}
  \caption{\textcolor{black}{A framework overview of {\it PoDiRe} method}}
  \label{fig:methods_overview}
\end{figure}

To this end, we propose {\it PoDiRe} to integrate MTL into an RL-based recommender. As shown in Fig.~\ref{fig:methods_overview}, {\it PoDiRe} consolidates multiple recommendation policies into a single multi-task policy via three steps. First, multiple distinct recommendation policies are learned in parallel, each of which handles one particular recommendation task. Each of such policies is encoded by a DNN referred to as a ``teacher network''. The input of a teacher network is the state and action representation, and the output is the estimated state-action value of taking an action at one state. Second, the training data of the consolidated multi-task policy is generated by calling the well-trained teacher networks obtained in the first step. For instance, given a collection of state-action pairs, it collects and stores the estimated state-action values for all recommendation tasks. Third, a multi-task recommendation policy is learned via a DNN by using the training data generated in the second step. The DNN  encodes the multi-task policy and is referred to as a ``student network''. The resulting student network is capable of estimating the state-action values in multiple recommendation tasks simultaneously. 

{\it PoDiRe} takes several measures to tackle the aforementioned challenges for integrating MTL with RL-based recommenders. First, to encourage knowledge sharing among different tasks in the third step, {\it PoDiRe} learns common feature representations among different tasks, and simultaneously minimizes the distance between the state-action values estimated by the student network and by the teacher networks. Second, to handle the issue that the scales of the state-action values in different tasks differ a lot, {\it PoDiRe} leverages a soft-max layer to transform the state-action values into a probability distribution and then computes the Kullback-Leibler (KL) divergence as a distance metric. Third, to reduce the risk of gradient interference in the exploration-and-error process, {\it PoDiRe} decouples the step of exploration-and-error and the step of training multi-task policy. The exploration-and-error process happens only in the training of teacher networks and is avoided in the training of student network. Besides, it is worth noting that, when generating training data for student network, some state-action pairs have never been observed in the training of teacher networks. However, the state-action values of these pairs can still be obtained by feeding the corresponding state and action into the well-trained teacher networks. This brings extra information for training student network and thus makes it possible to further improve the performance on individual tasks while reducing the size of the model~\cite{bucilua2006model,tang2018ranking}. 

Section~\ref{subsec:teacher} will introduce the developed teacher network and its training algorithm, where the DDQN is used to encode the teacher network. The technical details of the student network will be discussed in Section~\ref{subsec:student_distill}. In Section~\ref{subsec:state_representation}, the state and action representation learning method for the student and teacher networks is presented.

%% file: 4_2_teacher_model.tex
\subsection{Teacher Model}\label{subsec:teacher}
To address the long-term rewards of recommendations, RL is used to derive the optimal recommendation policy, which corresponds to the optimal state-action value. A state-action value under a policy is defined as the expected long-term rewards when taking an action at one state and following the policy thereafter. To be more specific, given an arbitrary policy $\pi(\cdot):\mathcal{S}\to\mathcal{A}$ that can generate action at any state $\bm{s}\in \mathcal{S}$ , the state-action value under the given policy can be expressed as: $Q^{(i)}_{\pi}(\cdot,\cdot):\mathcal{S}\times\mathcal{A}\to\mathbb{R}$ in the $i$-th task, i.e.,
\begin{align}
    Q^{(i)}_{\pi}(\bm{s},\bm{a}) := \mathbb{E}_{\mathcal{P}^{(i)}}\big[r^{(i)}_t + \gamma r^{(i)}_{t+1} + \gamma^{2} r^{(i)}_{t+2} + \cdots |\bm{s}_t=\bm{s},\bm{a}^{(i)}_t=\bm{a}, \pi\big],
\end{align}
where the expectation is taken over the probability distribution of state transition $\mathcal{P}^{(i)}$ and $\gamma \in [0, 1]$ is the discount factor for future rewards, e.g., $\gamma = 0$ means that the recommender considers only immediate rewards and $\gamma = 1$ indicates that it treats future rewards and immediate rewards equally. Note that although different tasks have shared state space $\mathcal{S}$ and the set of actions $\mathcal{A}$, the probability distribution of state transition $\mathcal{P}^{(i)}$ and the immediate rewards at $t$-th recommendation $r_{t}^{(i)}$ may differ from task to task. As a result, the state-action value $Q^{(i)}_{\pi}(\bm{s},\bm{a})$ would also be different for different $i$. Let $Q^{(i)}(\bm{s},\bm{a})$ denote the optimal state-action value over all policies for the $i$-th task, that is:
\begin{align}
    Q^{(i)}(\bm{s},\bm{a}) := \argmax_{\pi}Q^{(i)}_{\pi}(\bm{s},\bm{a}).
\end{align}
The optimal state-action value corresponds to the optimal policy. Let $\pi_{T}^{(i)}(\cdot)$ denote the optimal policy for the $i$-th recommendation task, where the subscript $T$ indicates a teacher policy. It can be derived from the optimal values by selecting the highest-valued action in each state:
\begin{align} \label{eq:pi_i}
    \pi_{T}^{(i)}(\bm{s}) := \argmax_{\bm{a}\in \mathcal{A}}Q^{(i)}(\bm{s}, \bm{a}).
\end{align}

The optimal state-action values can be estimated for each state-action pair using $Q$-learning~\cite{watkins1989learning}. In recommendation tasks, due to the large number of users and items to consider, the space of state-action pairs can be prohibitively large to learn. Instead, we can learn a parametric value function $Q^{(i)}(\cdot, \cdot;\Theta^{(i)})$ encoded by a DNN with parameters $\Theta^{(i)}$ ($\Theta^{(i)}$ will be discussed together with the state and action representation in Section~\ref{subsec:state_representation}). Let $y_{t}^{(i)}$ denote the target value of $Q^{(i)}(\bm{s}_t, \bm{a}^{(i)}_t;\Theta^{(i)})$. The long-term rewards of taking action $\bm{a}^{(i)}_{t}$ at state $\bm{s}_{t}$ are composed of two parts: the immediate reward $r_{t}^{(i)}$ and the future reward obtained by following the optimal policy $\pi_{T}^{(i)}(\cdot)$ at the transited state $\bm{s}_{t+1}$. As such, $y_{t}^{(i)}$ can be determined by:
\begin{align}\label{eq:q_i}
    y_{t}^{(i)} = r_{t}^{(i)} + \gamma \max_{\bm{a}\in\mathcal{A}}Q^{(i)}\big(\bm{s}_{t+1}, \bm{a};\Theta^{(i)}\big).
\end{align}
Note that in Equation (\ref{eq:q_i}), $Q^{(i)}(\cdot, \cdot; \Theta^{(i)})$ is used to select an action (i.e., computing the optimal action at state $\bm{s}_{t+1}$) and evaluate the quality of the action (i.e., computing the target value of $Q^{(i)}(\bm{s}_t,\bm{a}^{(i)}_{t}; \Theta^{(i)})$. As a consequence, it is more likely to select over-estimated values and result in over-optimistic value estimates~\cite{mnih2015human}. To alleviate this issue, van Hasselt \textit{et al}.~\cite{hasselt2016deep} propose the Double Deep Q-Network (DDQN), which has been successfully applied to solve the single-task recommendation problem~\cite{chen2018}.

\begin{algorithm}[t]
  \BlankLine
  \KwIn{$N_r$: Replay buffer maximum size; \\
  $N_b$: Training batch size; \\
  $T^{(i)}_{-}$: The time steps between two updates of the target network; \\
  $\eta^{(i)}$: Learning rate; \\
  $T_e$: Total number of epochs.\\
  }
  \KwOut{$\Theta^{(i)}$}
  Initialize $\Theta^{(i)} = \Theta^{(i)}_{-} = \Theta^{(i)}$ with random weights;\\
  Initialize replay buffer $\mathcal{B}=\emptyset$; \\
    \For {$epoch = 1, \cdots, T_{e} $}{
        \For {$j = 1, \cdots, t$}{
            Given any user, obtain current state $\bm{s}_{j}$;\\
            With probability $\epsilon$, recommend a random item represented by $\bm{a}^{(i)}_{j}$;\\
            With probability $1-\epsilon$, recommend the item $\bm{a}^{(i)}_j = \argmax_{\bm{a}\in\mathcal{A}}Q^{(i)}(\bm{s}_j, \bm{a};\Theta^{(i)})$;\\
            Observe and transform feedback into reward signal $r_{j}^{(i)}$;\\
            Observe and obtain user's next state $\bm{s}_{j+1}$; \\
            \If{$|\mathcal{B}|>N_{r}$}{
                Remove the oldest in $\mathcal{B}$;
            }
            Store $(\bm{s}_{j}, \bm{a}^{(i)}_{j}, \bm{s}_{j+1}, r_{j}^{(i)})$ in $\mathcal{B}$;\\
            Randomly sample $N_{b}$ records in $\mathcal{B}$;\\
            Compute $y^{(i)}_{j} = r^{(i)}_{j} + \gamma Q^{(i)} \big(\bm{s}_{j+1},\argmax_{\bm{a}\in\mathcal{A}}Q^{(i)}(\bm{s}_{j+1}, \bm{a}; \Theta^{(i)}); \Theta^{(i)}_{-}\big)$ for each record;\\
            Perform a gradient descent for minimizing $\mathcal{L}(\Theta^{(i)})=\frac{1}{2N_{b}}\sum\big(y^{(i)}_{j}-Q^{(i)}(\bm{s}_{j}, \bm{a}^{(i)}_{j}; \Theta^{(i)})\big)^{2}$;\\
            $\Theta^{(i)}\leftarrow \Theta^{(i)} - {\eta^{(i)}}\nabla_{\Theta^{(i)}}\mathcal{L}(\Theta^{(i)})$;\\
            \If{$j \  mod \ T_{-} = 0$}{
                $\Theta^{(i)}_{-} = \Theta^{(i)}$;
            }
        }
    }
  \Return{ $\Theta_{(i)}$; 
  }
\caption{Training algorithm of the $i$-th teacher model}
\label{alg:teacher}
\end{algorithm}

In DDQN, two networks are used: the current network $Q^{(i)}(\cdot,\cdot; \Theta^{(i)})$ to select an action and another target network represented by $Q^{(i)}(\cdot,\cdot; \Theta_{-}^{(i)})$ to evaluate the action. It updates the parameters of the target network with the parameters of the current network every $T_{-}$ time steps. Here $\Theta^{(i)}_{-}$ denotes the parameters of the target network and $T_{-}$ denotes the time steps between two updates of the target network. Then we can rewrite the target value of $Q(\bm{s}_t,\bm{a}_t;\Theta^{(i)})$ as:
\begin{align}
    y^{(i)}_{t} = r^{(i)}_{t} + \gamma Q^{(i)} \big(\bm{s}_{t+1},\argmax_{\bm{a}\in\mathcal{A}}Q^{(i)}(\bm{s}_{t+1}, \bm{a}; \Theta^{(i)}); \Theta^{(i)}_{-}\big).
\end{align}
Since $\Theta^{(i)}$ is trained to minimize the difference between the target value and the $Q$-value, the loss function of $\Theta^{(i)}$, denoted by $\mathcal{L}(\Theta^{(i)})$, can be written as:
\begin{align}
    \mathcal{L}(\Theta^{(i)}) := \mathbb{E}_{\bm{s}_t,\bm{a}_t, r^{(i)}_t, \bm{s}_{t+1}\sim u(\mathcal{B})} \Big[\frac{1}{2}\big(y^{(i)}_{t} - Q^{(i)}(\bm{s}_{t}, \bm{a}^{(i)}_{t}; \Theta^{(i)})\big)^{2}\Big],
\end{align}
where $\mathcal{B}$ denotes the replay buffer providing a sampling pool for batch updates and $u(\mathcal{B})$ denotes the uniform distribution over the instances in replay buffer $\mathcal{B}$. To be more specific, $\mathcal{B}=\{e_{k}\}_{i=k}^{N_r}$, and $e_{k} = (\bm{s}_k,\bm{a}_k, r^{(i)}_k, \bm{s}_{k+1})$. $N_r$ is the capacity of the replay buffer. In each batch update of the parameters $\Theta^{(i)}$, instead of directly using the current instance, that will make two consecutive batch highly correlated, a batch of instances are uniformly sampled from replay buffer $\mathcal{B}$. This can reduce the correlation among the training instances in the batch as well as the variance of the model~\cite{mnih2015human}.
Then the gradient of $\mathcal{L}(\Theta^{(i)})$ can be computed as:
\begin{align}
\nabla_{\Theta^{(i)}}\mathcal{L}(\Theta^{(i)}) = \mathbb{E}_{\bm{s}_t,\bm{a}_t, r^{(i)}_t, \bm{s}_{t+1}\sim u(\mathcal{B})} \Big[-\big(y^{(i)}_{t} - Q^{(i)}(\bm{s}_{t}, \bm{a}^{(i)}_{t}; \Theta^{(i)})\big)\nabla_{\Theta^{(i)}}Q^{(i)}(\bm{s}_{t}, \bm{a}^{(i)}_{t}; \Theta^{(i)})\Big].
\end{align}

The details of the training algorithm for the $i$-th teacher model are given in Algorithm~\ref{alg:teacher}, which is applicable to any $i\in\{1,2,\cdots,N_f\}$. Lines 6-7 correspond to the exploration-and-error process. Note that in this subsection, we assume the existence of state $\bm{s}_t$ and action $\bm{a}^{(i)}_t$ without explaining how to obtain them. The details about the representations of state and action will be provided in Section~\ref{subsec:state_representation}, where we will introduce how the state and action representations are jointly learned with $\Theta^{(i)}$.

%% file: 4_3_student_model.tex
\subsection{Student Model}\label{subsec:student_distill}
\begin{figure}[t]
  \centering
  \includegraphics[width=0.95\linewidth]{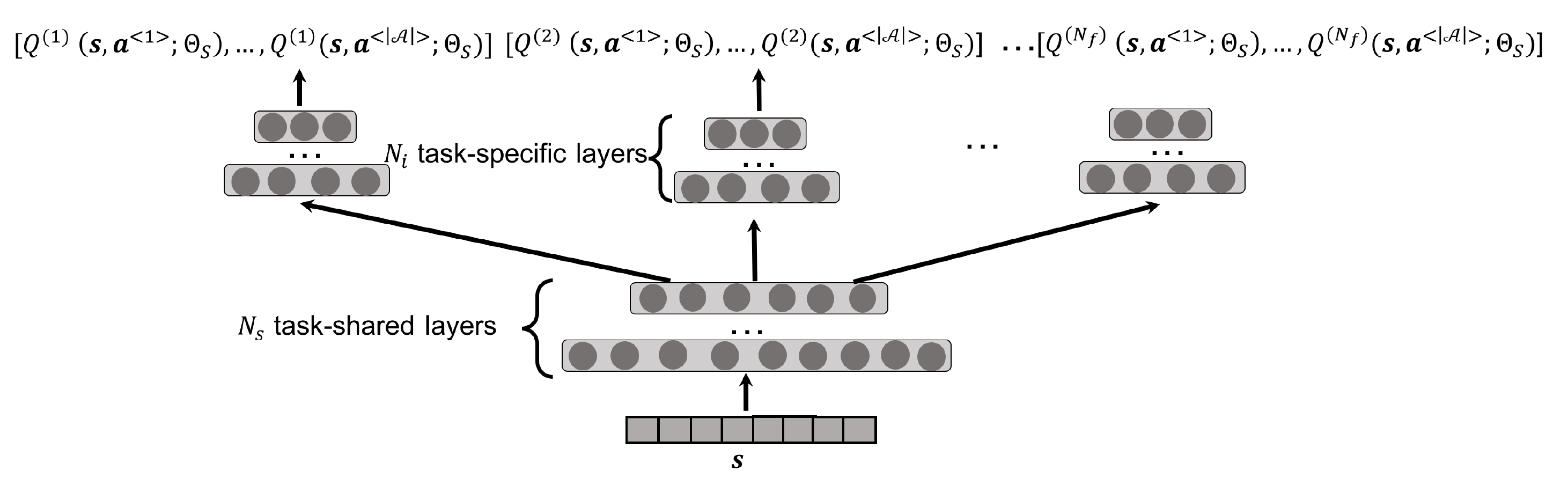}
  \caption{A DNN architecture of student network}
  \label{fig:student}
  \vspace{-2mm}
\end{figure}

When all teacher models are trained (i.e., $\{\Theta^{(i)}\}_{i=1}^{N_f}$ are all available), the $N_f$ recommendation tasks can be handled separately by the corresponding teacher. However, separately handling $N_f$ tasks cannot leverage knowledge sharing among different tasks, which is critical for improving the performance of individual tasks. To overcome this limitation, we propose to train a multi-task student policy $\bm{\pi}_{S}:=\{\pi^{(i)}_{S}(\cdot)\}_{i=1}^{N_f}$, whose $i$-th part $\pi^{(i)}_{S}(\cdot)$ mimics the $i$-th teacher for handling the $i$-th recommendation task. Supposing that $\bm{\pi}_{S}$ is parameterized by a DNN, the knowledge sharing among different tasks is realized by forcing all parts of $\bm{\pi}_{S}$ to share a common feature representation and \emph{simultaneously} minimizing their distance to the corresponding teacher policy. Fig.~\ref{fig:student} illustrates the architecture of the student network that is encoded by a feedforward DNN, where the input is a state $\bm{s}_{t}$ and the output has $N_f$ branches. Each one is a $Q^{(i)}(\cdot,\cdot)$-value vector for some $i$, from left to right, corresponding to the recommendation task $i = 1, 2, \cdots, N_f$. The $i$-th branch has $N_{i}$ task-specific layers and captures the distinct part of $\pi^{(i)}_{S}(\cdot)$. Meanwhile, all the branches share the first $N_{s}$ layers at the bottom to facilitate knowledge sharing among different tasks.

The output of the $i$-th branch is a $Q^{(i)}(\cdot,\cdot)$-value vector. For $\pi^{(i)}_{S}(\cdot)$ that is to mimic $\pi^{(i)}_{T}(\cdot)$, it is expected that the actions taken based on the $Q^{(i)}(\cdot,\cdot)$ values by the $i$-th branch should be close to those based on the output by the $i$-th teacher in the same state. A straightforward solution would be minimizing the distance between the outputs by $i$-th part of the student and the $i$-th teacher for any $i=1,2,\cdots,N_f$. Unfortunately, as explained before, for the same state-action pair, the estimated $Q^{(i)}(\cdot,\cdot)$ value is unbounded and may differ substantially among $i=1,2,\cdots,N_f$. Then the loss in minimizing one branch can be much larger than that for another branch, dominating the total loss in training. Moreover, the same action may be driven by more than one set of effective $Q^{(i)}(\cdot,\cdot)$ values as long as the relative rankings of actions based on their $Q^{(i)}(\cdot,\cdot)$ values is similar. For example, assume that at the same state there are three actions in the action space. Let the $Q^{(i)}(\cdot,\cdot)$ values of one policy be $(1, 2, 3)$ and those of another be $(10, 20, 30)$. Although the $Q^{(i)}(\cdot,\cdot)$ values are quite different, they lead to a similar tendency in choosing actions, i.e., the preference of $\bm{a}^{<3>}$ over $\bm{a}^{<2>}$ and $\bm{a}^{<1>}$. 

To address the above challenge, {\it PoDiRe} first uses softmax to transform the $Q^{(i)}(\cdot,\cdot)$ values into a probability distribution and then minimizes the KL divergence between the two distributions. After the softmax transformation, the unbounded $Q^{(i)}(\bm{s},\bm{a})$ values are set between $0$ and $1$ for each state-action pair $(\bm{s},\bm{a})$ without changing its relative ranking compared to other state-action pairs. This transformation is smooth and differentiable, which makes it easy to derive the gradient for parameter inference. Specifically, at some state $\bm{s} \in \mathcal{S}$, the approximated $Q^{(i)}(\cdot,\cdot)$-value vector from the $i$-th teacher's DDQN after transformation, denoted by $\bm{q}_{\tau}^{(i)}(\bm{s};\Theta^{(i)}) \in \mathbb{R}^{1\times |\mathcal{A}|}$, is
\begin{align}
    \bm{q}^{(i)}_{\tau}(\bm{s};\Theta^{(i)}) := \Big(S_{\tau}\big(Q^{(i)}(\bm{s},\bm{a}^{<1>}; \Theta^{(i)})\big),\cdots,S_{\tau}\big(Q^{(i)}(\bm{s},\bm{a}^{<|\mathcal{A}|>}; \Theta^{(i)}\big)\Big), 
\end{align}
where $S_{\tau}(\cdot)$ is softmax function with temperature $\tau > 0$:
\begin{align}
    S_{\tau}\big(Q^{(i)}(\bm{s},\bm{a}^{<k>}; \Theta^{(i)})\big):=\frac{\exp(Q^{(i)}(\bm{s},\bm{a}^{<k>}; \Theta^{(i)})/\tau)}{\sum_{l=1}^{|\mathcal{A}|}\exp(Q^{(i)}(\bm{s},\bm{a}^{<l>}; \Theta^{(i)})/\tau)}\hspace{4pt}.
\end{align}
The temperature $\tau$ controls how much knowledge is transferred from a teacher to the student. Raising the temperature will soften the transformed probability distribution and allow more knowledge to be transferred to the student~\cite{hinton2015distilling}. Let $\Theta_S$ denote the set of unknown parameters including weights and biases for all layers of the student network. We then represent the transformed $Q^{(i)}(\cdot,\cdot)$-value vector of the student model by $\bm{q}_{\tau}^{(i)}(\bm{s};\Theta_S)$:
\begin{align}
    \bm{q}^{(i)}_{\tau}(\bm{s};\Theta_S) := \Big(S_{\tau}\big(Q^{(i)}(\bm{s},\bm{a}^{<1>}; \Theta_S)\big),\cdots,S_{\tau}\big(Q^{(i)}(\bm{s},\bm{a}^{<|\mathcal{A}|>}; \Theta_S\big)\Big).
\end{align}
Let $D_{KL}(\cdot || \cdot)$ be the KL divergence function. Then the similarity between $\pi^{(i)}_{S}(\cdot)$ and $\pi^{(i)}_{T}(\cdot)$ may increase by minimizing the loss function given below:
\begin{align}
    \mathcal{L}(\Theta_S|\Theta^{(i)}):=  \mathbb{E}\Big[D_{KL}\big(\bm{q}_{\tau}^{(i)}(\bm{s}; \Theta^{(i)})||\bm{q}^{(i)}_{\tau}(\bm{s};\Theta_{S})\big)\Big].\label{equation: sub loss function}
\end{align}
Since as illustrated by Equation (\ref{eq:pi_i}), $\pi^{(i)}_{S}(\cdot)$ and $\pi^{(i)}_{T}(\cdot)$ can be derived by selecting the highest-valued action in each state, minimizing the KL divergence of $\bm{q}_{\tau}^{(i)}(\bm{s}; \Theta^{(i)})$ and $\bm{q}^{(i)}_{\tau}(\bm{s};\Theta_{S})$ is equivalent to minimizing the distance of $\pi^{(i)}_{T}(\cdot)$ and $\pi^{(i)}_{S}(\cdot)$. During this process, the knowledge of handling the $i$-th task forced to be distilled into the a part of the student from the $i$-th teacher.

Minimizing the loss function $\mathcal{L}(\Theta_S|\Theta^{(i)})$ encourages knowledge transfer between the student and the $i$-th teacher. To facilitate knowledge sharing among different tasks, {\it PoDiRe} proposes to \emph{simultaneously} minimize the summation of the loss between the student and multiple teachers, instead of a single loss. To be more specific, the loss function to navigate the learning of $\Theta_S$ between the student and teacher models can be derived as:
    \begin{equation}
        \mathcal{L}(\Theta_S|\{\Theta^{(i)}\}_{i=1}^{N_f}):=\sum_{i=1}^{N_f} \lambda_i \mathcal{L}(\Theta_S|\Theta^{(i)}),\label{equation:loss function}
    \end{equation}
where $\lambda_i$ denotes the weight of the loss between the student and the $i$-th teacher and can be determined through cross-validation. 
Then the gradient of $\mathcal{L}(\Theta_S|\{\Theta^{(i)}\}_{i=1}^{N_f}):=\sum_{i=1}^{N_f} \lambda_i \mathcal{L}(\Theta_S|\Theta^{(i)})$ with respect to $\Theta_S$ can be derived as
\begin{align}
    \nabla \mathcal{L}(\Theta_S|\{\Theta^{(i)}\}_{i=1}^{N_f})&=\sum_{i=1}^{N_f} \lambda_i \mathbb{E}\bigg[\sum_{\bm{a}\in\mathcal{A}}S_{\tau}\big(Q^{(i)}(\bm{s},\bm{a}; \Theta^{(i)})\big)\cdot \nabla \log S_{\tau}\big(Q^{(i)}(\bm{s},\bm{a}; \Theta_S)\big)\bigg]\\
    &=\sum_{i=1}^{N_f} \lambda_i \mathbb{E}\bigg[\sum_{\bm{a}\in\mathcal{A}}\frac{S_{\tau}\big(Q^{(i)}(\bm{s},\bm{a}; \Theta^{(i)})\big)}{S_{\tau}\big(Q^{(i)}(\bm{s},\bm{a}; \Theta_S)\big)}\cdot \nabla S_{\tau}\big(Q^{(i)}(\bm{s},\bm{a}; \Theta_S)\big)\bigg].\label{eq:true gradient}
\end{align}
Based on Equation (\ref{eq:true gradient}), given a batch of samples $\{(\bm{s}_{j}, \{\bm{q}_{\tau}^{(i)}(\bm{s}_j;\Theta^{(i)})\}_{i=1}^{N_{f}})\}_{j=1}^{N_b}$, where $N_b$ is the batch size, we can instantiate the stochastic gradient descent update for $\mathcal{L}(\Theta_S|\{\Theta^{(i)}\}_{i=1}^{N_f})$ as:
\begin{align}
    \Theta_S\leftarrow \Theta_S - \eta_s\cdot \frac{1}{N_b}\sum_{j=1}^{N_b}\sum_{i=1}^{N_f}\lambda_i\bigg( \sum_{\bm{a}\in\mathcal{A}}\frac{S_{\tau}\big(Q^{(i)}(\bm{s}_j,\bm{a}; \Theta^{(i)})\big)}{S_{\tau}\big(Q^{(i)}(\bm{s}_j,\bm{a}; \Theta_S)\big)}\cdot \nabla S_{\tau}\big(Q^{(i)}(\bm{s}_j,\bm{a}; \Theta_S)\big)\bigg).
    \label{eq:gradient}
\end{align}

\begin{algorithm}[t]
  \KwIn{Collections of $\{(\bm{s}_{t}, \{\bm{q}_{\tau}^{(i)}(\bm{s}_t;\Theta^{(i)})\}_{i=1}^{N_{f}})\}_{t=1}^{N}$}
  \KwOut{$\Theta_{S}$ for the student network}
  \BlankLine
  { Initialize $\Theta_{S}$ with random weights; \\
  }
  \For{ $epoch =1, \cdots, T_e$}{
  Draw a batch of $N_b$ training samples $\big\{(\bm{s}_j,\bm{q}_{\tau}^{(i)}(\bm{s}_j; \Theta^{(i)})\big\}_{j=1}^{N_b}$;\\
  $\Theta_S\leftarrow \Theta_S - \eta_s \frac{1}{N_b}\sum_{j=1}^{N_b}\sum_{i=1}^{N_f}\lambda_i\Big( \sum_{\bm{a}\in\mathcal{A}}\frac{S_{\tau}\big(Q^{(i)}(\bm{s}_j,\bm{a}; \Theta^{(i)})\big)}{S_{\tau}\big(Q^{(i)}(\bm{s}_j,\bm{a}; \Theta_S)\big)}\cdot \nabla S_{\tau}\big(Q^{(i)}(\bm{s}_j,\bm{a}; \Theta_S)\big)\Big)$;\label{alglist:theta update}
  }
  \Return{ $\Theta_{S}$;
  }
\caption{Training algorithm of student model}\label{alg:student}
\end{algorithm}

Note that optimizing $\mathcal{L}(\Theta_S|\{\Theta^{(i)}\}_{i=1}^{N_f})$ is equivalent to simultaneously minimizing the distance (in the policy space) between one part of the student policy and the teacher policy it aims to mimic. During this process, the knowledge from different teacher policies is forced to be jointly learned by the student policy. Through making use of the relatedness of different knowledge, the student policy is expected to attain better performance than each teacher in an individual recommendation task. To promote this knowledge-sharing process, supervised learning usually introduces a regularization term into the loss function~\cite{evgeniou2004regularized}. The regularization term will penalize the learning process if the parameter values of different tasks are far from some shared value, where the shared value can be some fixed value or the mean value over all tasks\footnote{We compared the performance of {\it PoDiRe} with this kind of approaches in Section \ref{section:experiment}.}. The loss function in {\it PoDiRe} $\mathcal{L}(\Theta_S|\{\Theta^{(i)}\}_{i=1}^{N_f})$ instead does a similar thing in policy space, as there is no correspondence between the set of parameters in student network and that in the teacher networks. The procedure of training the student model is presented in Algorithm~\ref{alg:student}.  

\begin{figure}[t]
  \centering
  \includegraphics[width=0.9\linewidth]{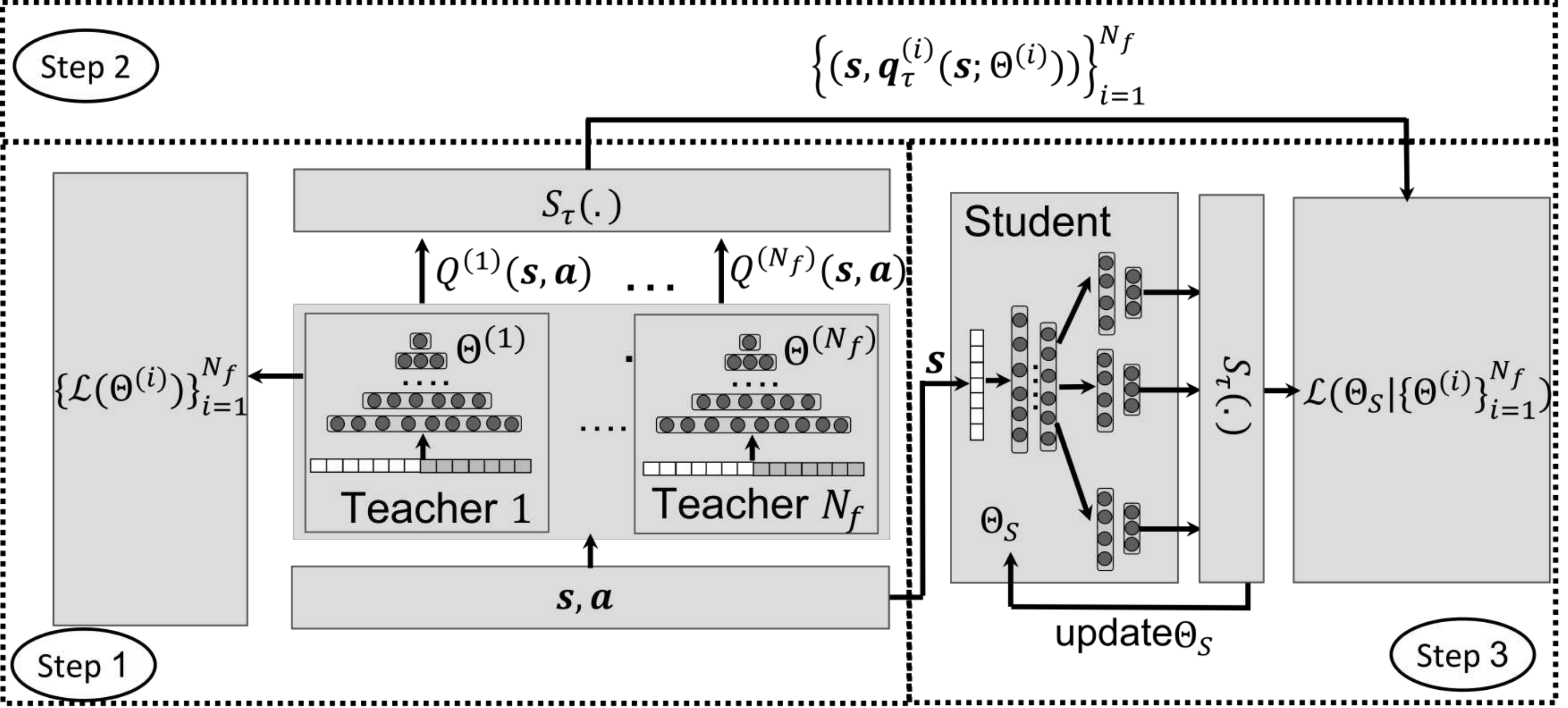}
  \vspace{-3mm}
  \caption{A detailed illustration of {\it PoDiRe} method}
  \label{fig:flowchart}
\end{figure}

Now, we can summarize the training procedures of {\it PoDiRe}. The detailed three steps are illustrated in Fig.~\ref{fig:flowchart}, \textcolor{black}{which correspond to the framework in Fig.~\ref{fig:methods_overview}}. The training is an iterative process with three steps in each iteration. The first step is to train the teacher models in parallel by following Algorithm~\ref{alg:teacher}. After all the teachers are trained, the set of their parameters $\{\Theta^{(i)}\}_{i=1}^{N_f}$ is obtained. These parameters are used in the second and third steps. In the second step, the learned teacher models are leveraged to generate training instances $\{(\bm{s}_{j}, \{\bm{q}_{\tau}^{(i)}(\bm{s}_j;\Theta^{(i)})\}_{i=1}^{N_{f}})\}_{j=1}^{t}$ for the student model. Since $\{\Theta^{(i)}\}_{i=1}^{N_f}$ are available, given an arbitrary state $\bm{s}$, we are able compute the value of $\bm{q}_{\tau}^{(i)}(\bm{s}; \Theta^{(i)})$ for any $i=1,2,\cdots,N_f$, no matter whether the state is observed or not during the training of the teacher models. As such, except for the states used to train the teacher models, we also have the freedom to sample a proportion of states from the state space $\mathcal{S}$, and feed them into the well-trained teacher models to collect their outputs. The motivation of doing so is to provide an informative estimation of the $Q^{(i)}(\cdot,\cdot)$ values on unvisited states. This has been shown to be useful for improving the performance of the student as well as reducing the size of the student model~\cite{bucilua2006model,tang2018ranking}. In the third step, the training instances generated in the second step are fed into Algorithm~\ref{alg:student} to train the student model. In general, the three steps iteratively optimize Part I (loss of each teacher model) and Part II (distillation loss) of the loss function $\mathcal{L}(\{\Theta^{(i)}\}_{i=1}^{N_f}, \Theta_S)$, as shown below:
\begin{align}\label{eq:summed loss}
    \min \ \mathcal{L}(\{\Theta^{(i)}\}_{i=1}^{N_f}, \Theta_S) :&=\underbrace{\sum_{i=1}^{N_f}\mathcal{L}(\Theta^{(i)})}_{\text{Part I}} + \underbrace{\vphantom{ \sum_{i=1}^{N_f}\mathcal{L}(\Theta^{(i)})} \mathcal{L}(\Theta_S | \{\Theta^{(i)}\}_{i=1}^{N_f})}_{\text{Part II}} \nonumber \\ 
    &=\sum_{i=1}^{N_f}\big[\mathcal{L}(\Theta^{(i)}) + \lambda_i \mathcal{L}(\Theta_S|\Theta^{(i)})\big].
\end{align}
Part II of Equation (\ref{eq:summed loss}) can also be interpreted as a regularization term as the teacher policy parameterized by $\Theta^{(i)}$ is regularized by a shared student policy parameterized by $\Theta_{S}$. 

As a final note, it is worth mentioning that a recent study on ranking distillation (RD)~\cite{tang2018ranking} takes advantage of a similar distillation technique for recommender systems. However, it focuses on a single-task recommendation task while {\it PoDiRe} aims to handle multiple recommendation tasks. Their difference in the loss function is notable: while there is only one KL divergence component in the distillation loss of RD, the distillation loss of {\it PoDiRe} is composed of multiple components obtained from all $\pi^{(i)}_{S}(\cdot)$ and $\pi^{(i)}_{T}(\cdot)$ pairs. 

%% file: 4_4_state_action_representation.tex
\subsection{Action and State Representation}
\label{subsec:state_representation}

In this subsection, we introduce the action and state representation method for the student and teacher networks. The representation of user state has been shown to play a critical role in achieving satisfactory performance in many RL-based recommenders~\cite{zhao2018,zhao2018drl,zheng2018drn,chen2018,wang2018supervised,theocharous2015}. In these methods, the state representation is usually obtained from the $T$ most recent recommendations the user has interacted with, where $T$ is a hyper-parameter determined by cross-validation. In the recommendation practice with Samsung Game Launcher, we noticed that the state represented in these ways might change abruptly in two consecutive training instances due to the dynamics of user behaviors, making the training of the model unstable. To resolve this issue, we add into the representation a relevant and relatively static part that summarizes the ``long-term interest'' of the user. Each feature of this part describes statistics from a longer time horizon, e.g., statistics in all historical interactions, app usage, and user profiles. Taking app genre as an example, one possible feature is the distribution of genres of historically used apps by the user. The part similar to previous methods is referred to as ``short-term interest''. In recommendations, the part of long-term interest changes much more slowly than the part of short-term interest. As empirically evaluated in Section~\ref{section:experiment}, this mixture of long-term and short-term interest in state representation stabilizes the learning process and improves the performance of the resulting model. Finally, to take advantage of the relevance between contexts (e.g., time and location) and user feedback, contexts of one recommendation are also included in the state as the third part.

On the other hand, the information heterogeneity in action (item) representation has not been carefully discussed in most previous studies on RL-based recommenders, although it commonly exists in real-world applications. Taking app items as an example, the type of information available for an app ranges from unstructured data such as image and text to structured data such as app profiles, and aggregated user feedback to the app. In e-commerce examples, a product also has its picture (image), customer reviews (text), profiles and overall history of being purchased as correspondence. There have been several existing methods to fuse the heterogeneous information in one representation~\cite{chang2015heterogeneous}. Inspired by these methods, we propose our method below.

\begin{figure*}[t]
\centering
\includegraphics[width=.95\textwidth, height=.45\textwidth]{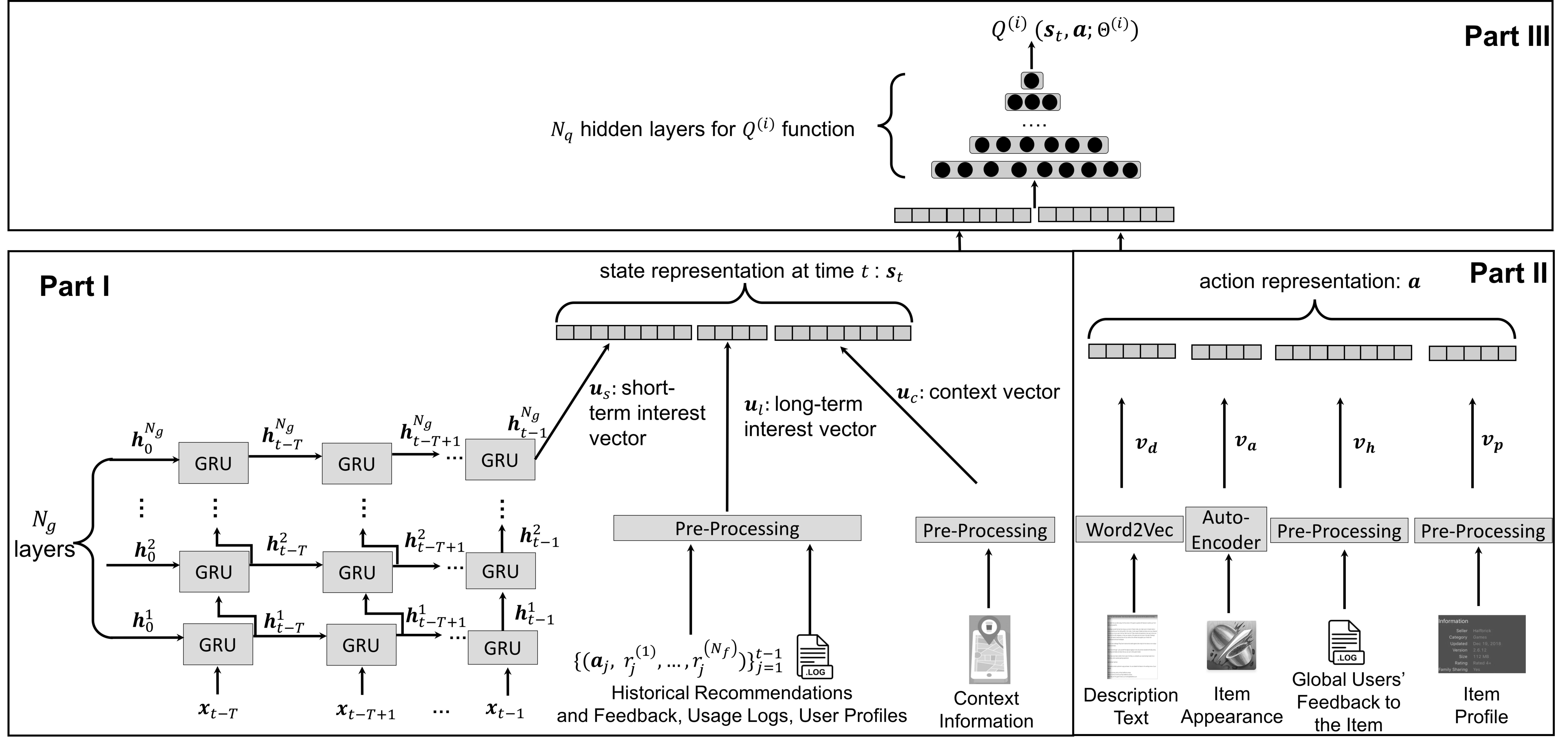} 
\vspace{-2mm}
\caption{\textcolor{black}{A DNN architecture of the $i$-th teacher network}}
\label{fig:teacher}
\vspace{-2mm}
\end{figure*}

\textbf{Action Representation.} Fig.~\ref{fig:teacher} illustrates a DNN architecture of the $i$-th teacher network. As illustrated in Part II of Fig.~\ref{fig:teacher}, several representation learning techniques are used to process the heterogeneous raw data of an item in order to embed them into a representation of the action. For example, the vector $\bm{v}_{d} \in \mathbb{R}^{1\times N_{d}}$ capturing the information in an item's textual description can be obtained by embedding the words using pre-trained Word2Vec~\cite{mikolov2013distributed}. The vector $\bm{v}_{a} \in \mathbb{R}^{1\times N_{a}}$ capturing the appearance of an item can be obtained through a pre-trained convolutional neural network (CNN) based auto-encoder~\cite{masci2011stacked}. Besides, the vector $\bm{v}_{h} \in \mathbb{R}^{1 \times N_{h}}$ representing the overall feedback to an item is obtained by aggregating the logs of all users' feedback, e.g., the weekly, bi-weekly, and monthly minimums, means, medians and maximums, of the overall historical feedback, etc. In addition, the vector $\bm{v}_{p} \in \mathbb{R}^{1 \times N_{p}}$ representing the profile of an item is obtained by parsing attributes like the maker and the category, etc. Let $N_{d}$, $N_{a}$, $N_{h}$ and $N_{p}$ denote the dimensions of feature vectors $\bm{v}_{d}$, $\bm{v}_{a}$, $\bm{v}_{h}$ and $\bm{v}_{p}$, respectively. The final representation of an action $\bm{a}$ is generated by concatenating the four parts: 
\begin{align}
    \bm{a} = \text{concat}(\bm{v}_{d}, \bm{v}_{a}, \bm{v}_{h}, \bm{v}_{p}).
\end{align}
 It is worth noting that the CNN-based auto-encoder and Word2Vec module are pre-trained with the textual description and images of items separately. In other words, their parameters are separately learned with $\Theta^{(i)}$. This is significantly different from the state representation introduced next, where the parameters of embedding layers are part of $\Theta^{(i)}$ and are jointly learned with parameters in other hidden layers. This brings benefits in directly using the item representations later in state representation.

\textbf{State Representation.} In our problem, a state is associated with a user at a particular time, and thus its representation is expected to capture the user's preference over time. It is very challenging to model such dynamics in state representation. A few recent studies~\cite{chen2019top,zhao2018,wang2018supervised,oosterhuis2018} have tackled this challenge by feeding into recurrent neural networks (RNNs) the representations of items that were interacted by users in $T$ most recent recommendations. This is referred to as short-term interest in {\it PoDiRe} as it usually reflects the most recent interests of the user. However, as mentioned earlier, such a solution may suffer from the abrupt change of states in two consecutive examples, which easily makes the training of the model unstable. Therefore, {\it PoDiRe} introduces into the state a relevant and relatively slowly-changing part along with the short-term interest part to achieve a trade-off.  As shown in Part I of Fig.~\ref{fig:teacher}, {\it PoDiRe} represents the state of a user at the $t$-th recommendation by three parts: the user's short-term interest $\bm{u}_{s}$ captured by some type of RNNs like in most previous studies, the user's long-term interest $\bm{u}_{\ell}$ obtained from the user's app usage logs, profiles as well as all historical recommendations and responses, and the context information $\bm{u}_{c}$ that contains the contextual information (e.g, location, time, etc) at the time of recommendation. We introduce each part in the following. 

{\tt Short-Term Interest Vector:} A user's short-term interest is usually reflected by his/her recent feedback to recommendations~\cite{chen2019top,zhao2018}. As such, a multi-layer gated recurrent unit (GRU) is introduced to capture the dynamics of a user's short-term interest\footnote{Both GRU and LSTM demonstrate better performance in handling vanishing gradients than vanilla RNN and can fulfill this task, but GRU is faster for training and more suitable for processing big data.}. The recommendations are fed into the GRU, each consisting of the representation of a recommended item and the user's feedback to the item. The hidden state of the multi-layer GRU is leveraged as the representation of the short-term interest vector $\bm{u}_{s}$. As illustrated by Part I of Fig.~\ref{fig:teacher}, let $\bm{x}_{t-j}$ be the input to the $(t-j)$-th GRU unit in the first layer. Then we have: 
\begin{align}
    \bm{x}_{t-j}:= \text{concat}(\bm{a}_{t-j}, r_{t-j}^{(1)},\cdots,r_{t-j}^{(N_{f})}),
\end{align}
where $j=1,2,\cdots,T$ and $T$ is the truncated time steps for GRU. Let $\bm{h}_{t-1}\in \mathbb{R}^{1\times N_{u}}$ denote the input hidden state to the ($t-1$)-th GRU unit, where $N_{u}$ is the dimension of the hidden state. Let $\bm{z}_{t-1} \in [0,1]^{1\times N_{u}}$ and $\bm{\widehat{h}}_{t-1}\in \mathbb{R}^{1\times N_{u}}$ denote the update gate and the proposed new hidden state, respectively. Let $\bm{z}_{t-1}$ denotes proportion of old hidden state $\bm{h}_{t-2}$ in representing the new hidden state $\bm{h}_{t-1}$. Let $\bm{r}_{t-1} \in \mathbb{R}^{1\times N_{u}}$ be the reset gate that moderates the impact of the old hidden state $\bm{h}_{t-2}$ on the new hidden state $\bm{\widehat{h}}_{t-1}$. Then we have:

\begin{align}
    \begin{cases}
        \bm{h}_{t-1} := \bm{z}_{t-1} \odot \bm{h}_{t-2} + (\bm{1}-\bm{z}_{t})\odot\bm{\widehat{h}}_{t-1}, \\ \nonumber
        \bm{z}_{t-1} := \sigma\big(\bm{x}_{t-1}\bm{W}_{xz}+\bm{h}_{t-2}\bm{W}_{hz} + \bm{b}_{z}\big), \\ \nonumber
        \bm{\widehat{h}}_{t-1} := \text{tanh}\big(\bm{x}_{t-1}\bm{W}_{xh} + (\bm{r}_{t-1}\odot\bm{h}_{t-2})\bm{W}_{hh} + \bm{b}_{h}\big), \\ \nonumber
        \bm{r}_{t-1} := \sigma\big(\bm{x}_{t-1}\bm{W}_{xr}+\bm{h}_{t-2}\bm{W}_{hr} + \bm{b}_{r}\big).
    \end{cases}
\end{align}
where $\odot$ is the Hadamard product, $\sigma(\cdot)$ is the sigmoid function, $\{\bm{W}_{xh}, \bm{W}_{hh}, \bm{b}_{h}\}$, $\{\bm{W}_{xz}, \bm{W}_{hz}, \bm{b}_{z}\}$, and $\{\bm{W}_{xr},\bm{W}_{hr},\bm{b}_{r}\}$ are the unknown weights and biases in representing the proposed new hidden state, the update gate, and the reset gate, respectively. The subscript $h$ corresponds to the hidden state, $z$ corresponds to the update gate,  and $x$ corresponds to the input vector $\bm{x}_{t-1}$. These weights and biases for the gates are shared across different GRUs in the same layer, and have a varying superscript in their notations. To differentiate them, we add superscript to them. For instance, $\bm{h}^{k}_{t-j}$ denotes the hidden state of the $k$-th layer and ($t-j$)-th GRU, where $k=1,2,\cdots,N_g$ with $N_g$ being the number of GRU layers. We use zero-mean small-variance Gaussian random noise to initialize $\{\bm{h}^{k}_{0}\}_{k=1}^{N_{g}}$, the initial state of the multi-layer GRU. Finally we have the short-term interest vector as:
\begin{align}
    \bm{u}_{s} := \bm{h}^{N_{g}}_{t-1}.
\end{align}

{\tt Long-Term Interest Vector:} The short-term interest of users might change much faster than long-term interest. If, as most previous studies did, only this part is used to represent the user state, the state of two consecutive recommendations might change abruptly, making the training of the model unstable. To mitigate the issue, we introduce a user's long-term interest vector $\bm{u}_{\ell}\in \mathbb{R}^{1\times N_{\ell}}$ into the state representation, where $\in \mathbb{R}^{1\times N_{\ell}}$ is the length of the vector. $\bm{u}_{\ell}$ changes more slowly than $\bm{u}_{s}$ among consecutive recommendations. \textcolor{black}{The features in $\bm{u}_{\ell}$ mainly come from a user's app usage data, user profile, as well as a user's historical interactions with recommendations but with an emphasis on long-term statistics. Particularly, the user's app usage information consists of the used apps' category distribution (causal, racing, sports, ...), price distribution (free, \$$1.99$, \$$2.99$, ...), rating distribution ($3$ stars or less, $4$ stars and $5$ stars), downloading distribution ($1,000$ downloads or less, $10,000$ downloads, $100,000$ downloads, ...), and etc. The user's profile includes age, gender, region, and etc. The user's historical interactions with recommendations could be the distributions of item categories that have obtained positive feedback from the user and the distributions of items' producers with positive feedback from the user.}

{\tt Context Vector:} Similarly, we encode important contextual attributes such as time and location into $\bm{u}_{c} \in \mathbb{R}^{1\times N_c}$, where $N_c$ is the length of $\bm{u}_{c}$. These types of information have been found useful to improve user experience in the cold-start scenarios in practice~\cite{Braunhofer2014}. Now we are ready to generate the overall representation of a user's state at the $t$-th recommendation:
\begin{align}\label{eq:state representation}
    \bm{s}_{t} := \text{concat}(\bm{u}_{s}, \bm{u}_{\ell}, \bm{u}_{c}),
\end{align}
where $\bm{s}_{t} \in \mathbb{R}^{1\times (N_{s}+N_{\ell}+N_{c})}$.

As shown in Part III of Fig.~\ref{fig:teacher}, given state and action representation, we use a $N_{q}$-layer feedforward neural network to parameterize the $Q^{(i)}(\cdot, \cdot)$, where $\text{ReLU}(\cdot)$ is the activation function for each layer. Let $\{\bm{W}_{q}^{k}, \bm{b}^{k}_{q}\}$ be the weights and bias of the $k$-th layer, where $k=1, 2,\cdots, N_q$. The parameters in the state and action representation and the parameters to approximate $Q^{(i)}(\cdot, \cdot)$ can be summarized in $\Theta^{(i)}$:
\begin{align}
    \Theta^{(i)} := \big\{\{\bm{W}_{q}^{k},\bm{b}_{q}^{k}\}_{k=1}^{N_q}, \{\bm{W}_{xz}^{k}, \bm{W}_{hz}^{k},\bm{W}_{xh}^{k}, \bm{W}_{hh}^{k},\bm{W}_{xr}^{k}, \bm{W}_{hr}^{k}, \bm{b}_{z}^{k}, \bm{b}_{h}^{k}, \bm{b}_{r}^{k}\}_{k=1}^{N_{g}}\big\}.
\end{align}

%% file: 5_experiments.tex
\section{Experiments}
\label{section:experiment}
Simultaneously operating multiple recommendation tasks with long-term rewards is an important business desideratum for many platforms to increase revenue and improve user experience. Unfortunately, to the best of our knowledge, all publicly available recommendation-related datasets are not suitable for evaluating the performance of {\it PoDiRe}. This is mainly because these datasets lack multiple-type feedbacks of recommendations, and the sequential dependency of those feedbacks that is captured in our model (i.e., {\it PoDiRe}). Therefore, in this paper, we collaborate with the Samsung Game Launcher platform, one of the largest commercial mobile gaming platforms in the world~\cite{SamsungGameLauncher}, to conduct comprehensive experiments for evaluating the performance of the {\it PoDiRe} method. The Samsung Game Launcher recommends mobile game apps to users and collects three types of sequentially-dependent feedbacks: click, install and play. Our recommendation goal is to simultaneously operate three recommendation tasks that maximize long-term clicks, installs, and plays, respectively. Thus, we have $N_f = 3$ in our experiments. Upon the internal approval of Samsung, we would like to release our data and source code to the public to facilitate future research in this area. We would like to note that our method (i.e., {\it PoDiRe}) is applicable to other applications as long as there are multiple recommendation tasks and long-term rewards to consider. 

In this section, the performance of the {\it PoDiRe} method is usually referred to as the test performance of the student network. The empirical evaluation of {\it PoDiRe} is composed of three parts. First, the {\it PoDiRe} method is evaluated in comparison with nine state-of-the-art methods in terms of both effectiveness and efficiency in Section~\ref{subsec: effectiveness} and Section~\ref{subsec: efficiency}. Details about these nine competing methods are provided in Section~\ref{sec:competingmethod}. The efficiency of the {\it PoDiRe} method is evaluated in terms of two aspects: the model size (number of parameters) after hyper-parameter tuning and response latency during the online test. Representatives of the baselines, as mentioned earlier, are compared with the {\it PoDiRe} method with respect to the two evaluation metrics. Second, to demonstrate the benefits of introducing long-term interest vector in the state representation, we conduct experiments to compare the performance of the Teacher Network with the long-term interest vector and without long-term interest vector in the state representation in Section~\ref{subsec:long-term interest}. In Section~\ref{subsec: different teachers}, we demonstrate the application of the proposed Teacher Network instead of other DRL based teacher models through comparing the performance of the student trained by those different teacher models.  Finally, in Section~\ref{subsec:effects of distillation}, we demonstrate the effectiveness of the knowledge-sharing through distillation by quantifying the performance improvement caused by the step of knowledge-sharing through distillation.

\subsection{Experimental Setup}
\begin{figure}[htbp]
  \centering
  \includegraphics[width=0.5\linewidth]{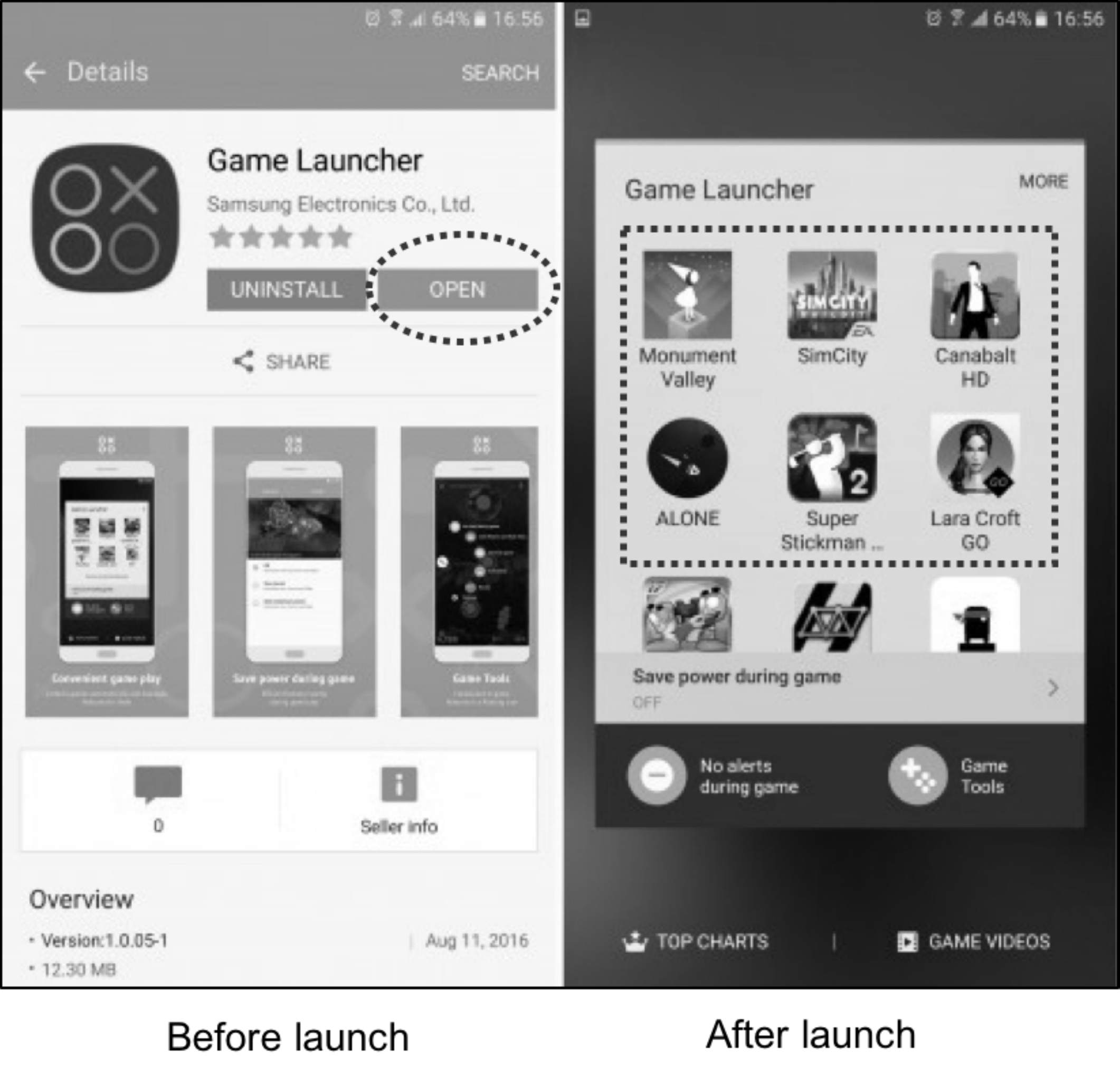}
  \vspace{-3mm}
  \caption{Interface of the Samsung Game Launcher recommender}
  \label{fig:gameLauncher}
\end{figure}
The experiments were conducted over a 5-week recommendation campaign from 10/20/2018 to 11/28/2018. The dataset contains a total of 2,483,321 recommendation sessions. The action set $\mathcal{A}$ contains 2,013 game apps. 3 types of feedbacks, click, install, and play, were collected together with the recommendation sessions. Fig.~\ref{fig:gameLauncher} illustrates the interface of the Samsung Game Launcher platform before launch and after launch. The dashed line before launch circled the button to open the game launcher platform, and examples of recommended game apps are shown inside the dashed rectangle after launch. The recommended game apps are displayed when a user launches the game platform, and are dynamically updated when the user consumes the recommendations. Each recommendation session contains the recommended games, user's multiple types of feedbacks (i.e., click, install, and play), and the timestamps of receiving the feedback. 


To generate the features of users' long-term interests for state representation and avoid data leakage, we collected and utilized another 4-week user-game interaction events before the campaign (from 9/20/2018 to 10/20/2018). The whole data set contains three major types of information: (1) play history, (2) game app profiles, and (3) user information. Each play record in the play history contains anonymous user id, a game package name, and the duration of the play. It is also accompanied with rich contextual information, such as WiFi connection status, screen brightness, audio volume, etc. Game profiles are collected from different game stores, including features like app icon, a textual description of contents, genre, developer, number of downloads, rating values, etc. As illustrated in Fig.~\ref{fig:teacher}, game profiles are also leveraged for learning action representation. User information contains the device model, region, OS version, etc. The key statistics of the data are summarized in Table~\ref{table:data_statistics}, where the number of impressions is the number after the aggregating by user-game pairs. 

\begin{table*}[htbp]
\centering
\caption{Dataset statistics}
\small
\begin{tabular}{|c|c|c|c|c|c|c|c|}
\hline
\textbf{Stage}   & \textbf{\# users} & \textbf{\# games} & \textbf{\# impressions} & \textbf{\# clicks} & \textbf{\# installs} & \textbf{\# plays} & \makecell{\textbf{\# events for} \\ \textbf{feature generation}} \\\hline
Training       & 477,348  &  2,013 & 1,173,590 & 330,892 & 29,556 & 18,642 & 620,633,212 \\ \hline
Testing    & 119,979  &  2,013 & 291,411 & 83,295 & 7,488 & 4,725 & 140,436,479 \\ \hline
\end{tabular}
\label{table:data_statistics}
\end{table*}

We split the collected recommendation sessions along time, and use 80\% as the training dataset and 20\% as the test dataset. The hyperparameters of the model, such as temperature $\tau$, learning rate $\eta$ and discount factor $\gamma$, etc., are tuned on a proportion of the training dataset. A grid search on these parameters was performed, and the combination yielding the best performance is chosen. \textcolor{black}{To facilitate the reproducibility, we enumerate the values of all parameters used for training the teacher models and student model as well as the baselines. We also report detailed software and hardware configurations. They are available in Section~\ref{subsec:parameter} and Section~\ref{subsec:configuration}, respectively.}

\subsection{Competing Methods}\label{sec:competingmethod}

We compare the {\it PoDiRe} method with nine baseline methods that can be grouped into three different categories: (1) traditional recommendation methods that are based on single-task supervised learning, (2) deep reinforcement learning methods that are based on single-task reinforcement learning, and (3) MTL methods that are based on multi-task supervised learning. The effectiveness of all competing methods is evaluated using $Precision@K$, $NDCG@K$, and $MAP$, which are the standard metrics used in previous research on recommender systems. The efficiency is evaluated using online response time as well as model size (i.e., the number of model parameters to learn). Specifically, the first category includes three methods: logistic regression (LR)~\cite{tang2016empirical}, factorization machines (FMs)~\cite{rendle2010factorization,rendle2012factorization}, and gradient boosting decision tree (GBDT)~\cite{wang2016mobile}. The three methods are based on supervised learning and rank the items to be displayed to a user based on the estimated probability that a user likes an item. The LR method estimates the probability through the logistic regression over the concatenated features of the user and the item. In LR, only the dependency between the output and the first-order features are investigated. To take into account second-order interaction between features, FMs learn an embedding for every single feature and model the interaction between two features through the dot product of their embeddings. LR and FMs assume a linear relationship between the features and output to be estimated. To overcome this limitation and further improve the performance in probability estimation, GBDT ensembles the decision tree as a nonlinear classifier to capture the non-linearity in the relationship between features and the output.

The second category includes DeepPage~\cite{zhao2018drl}, DRN~\cite{zheng2018drn}, and the Teacher Network proposed by this paper. The three methods are all based on deep reinforcement learning. Different from the methods in the first category, DRL based methods rank the items to be recommended based on their estimated long-term rewards once being liked by the user. Meanwhile, these methods assume that the long-term rewards depend on the state of the user, and the current recommendation may incur state transition of the user. Their difference is reflected in the aspect of state representation. DRN represents the state of a user mainly through the items that the user clicked in 1 hour, 6 hours, 24 hours, 1 week, and 1 year respectively. Different from DRN, DeepPage mainly uses the hidden state of an RNN and uses the most recent $T$ items browsed by the user as the input to the RNN. In the recommendation practice with Samsung Game Launcher, we noticed that the state represented in these ways might change abruptly in two consecutive training instances due to the dynamics of user behaviors, making the training of the model unstable. To resolve this issue, we add to the representation a relevant and relatively static part that summarizes the long-term interest of the user. Each feature of this part describes statistics from a longer time horizon, e.g., statistics in all historical interactions, app usage, and user profiles. In recommendations, the part of long-term interest changes much more slowly than the part of short-term interest. Note that these methods in the first two categories were designed to handle a single recommendation task. Thus they are unable to take advantage of of the knowledge sharing among different tasks to improve the performance on an individual task. Besides, to compare their performance with that of {\it PoDiRe}, we need to repeatedly train each baseline model $N_f$ times, each time being trained to handle one task. 

The third category includes three MTL-based methods: Sparse MTL (SMTL)~\cite{argyriou2007multi}, Feature-selected MTL (FMTL)~\cite{zhang2019multi}, and Regularized MTL (RMTL)~\cite{evgeniou2004regularized}. The DNNs for these methods are the same: the first 2 layers are shared, and then the shared layers are connected to three task-specific branches, each of which has a task-specific logistic regression layer and outputs a predicted value for one task. The input to the DNNs is the concatenated features of a user and an item. The output of one branch is the estimated probability that the user likes the item in some task. The loss function of these methods is composed of the cross-entropy loss plus an additional regularization term to encourage knowledge sharing among tasks. The cross-entropy loss is the same for these methods. The difference among these methods is their additional regularization terms, which reflect the ways to encourage knowledge-sharing among different tasks. 
Specifically, considering that the matrix $\bm{W}=[\bm{w}^{(1)},\bm{w}^{(2)},...,\bm{w}^{(N_f)}]$, where $\bm{w}^{(i)}$ is a vector is is the weights of the output layer for the $i-$th branch, then the additional regularization term of SMTL is formulated as $\mathcal{L}_{SMTL}(\bm{W}) := \|\bm{W}^{T}\|_{2,1}$, where $\|\cdot\|_{p,q}$ represents the $\ell_{p,q}$ norm for matrix. Since the row of $\bm{W}$ corresponds to a feature and a column of it represents an individual task, SMTL intends to rule out the unrelated features across tasks by shrinking the entire rows of the matrix to zero~\cite{wang2014multiplicative}. The additional regularization term of FMTL is $\mathcal{L}_{FMTL}(\bm{W}) := \|\bm{W}^{T}\|_{2,1}-\|\bm{W}^{T}\|_{2,2}$, where the first term achieves the group sparsity and the second term helps to learn task-specific features. The additional regularization term of RMTL is formulated as the distance between all task parameters to a set of shared parameters as: $\mathcal{L}_{RMTL}(\bm{W}) := \sum_{i=1}^{N_f}\|\bm{w}^{(i)} - 1/N_f\sum_{i=1}^{N_f}\bm{w}^{(i)}\|^{2}_{2}$, where $\|\cdot\|_{p}$ denotes the $\ell_p$ norm. RMTL assumes that the weights of all tasks are close to each other and thus penalizes the learning if the learned values fail to support this assumption. Note that the methods in the third category only need to be trained once to handle three tasks. They take into account the potential of knowledge-sharing in improving the performance over individual recommendation tasks. Compared to the {\it PoDiRe} method, the methods in the third category fail to consider the long-term rewards in ranking items to be recommended. These nine baseline methods provide a great representation for the state of the art.

\subsection{Effectiveness of {\it PoDiRe}}\label{subsec: effectiveness}

\begin{figure*}[htpb]
\centering
\subfigure[Click]{
\includegraphics[width=0.31\textwidth]{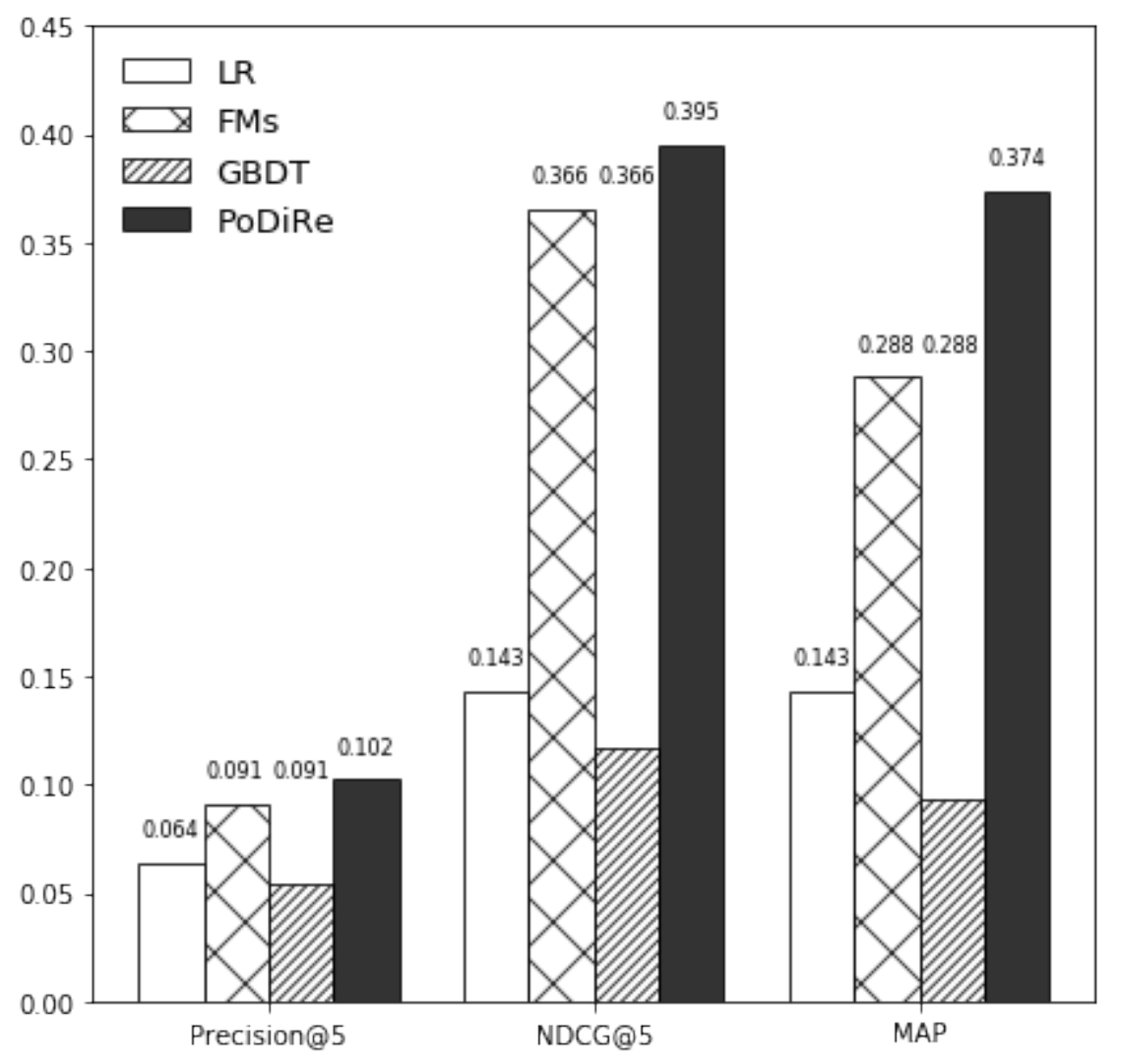}
\label{fig:comparison with single-task supervised learning - click}}
\hfill
\subfigure[Install]{
\includegraphics[width=0.31\textwidth]{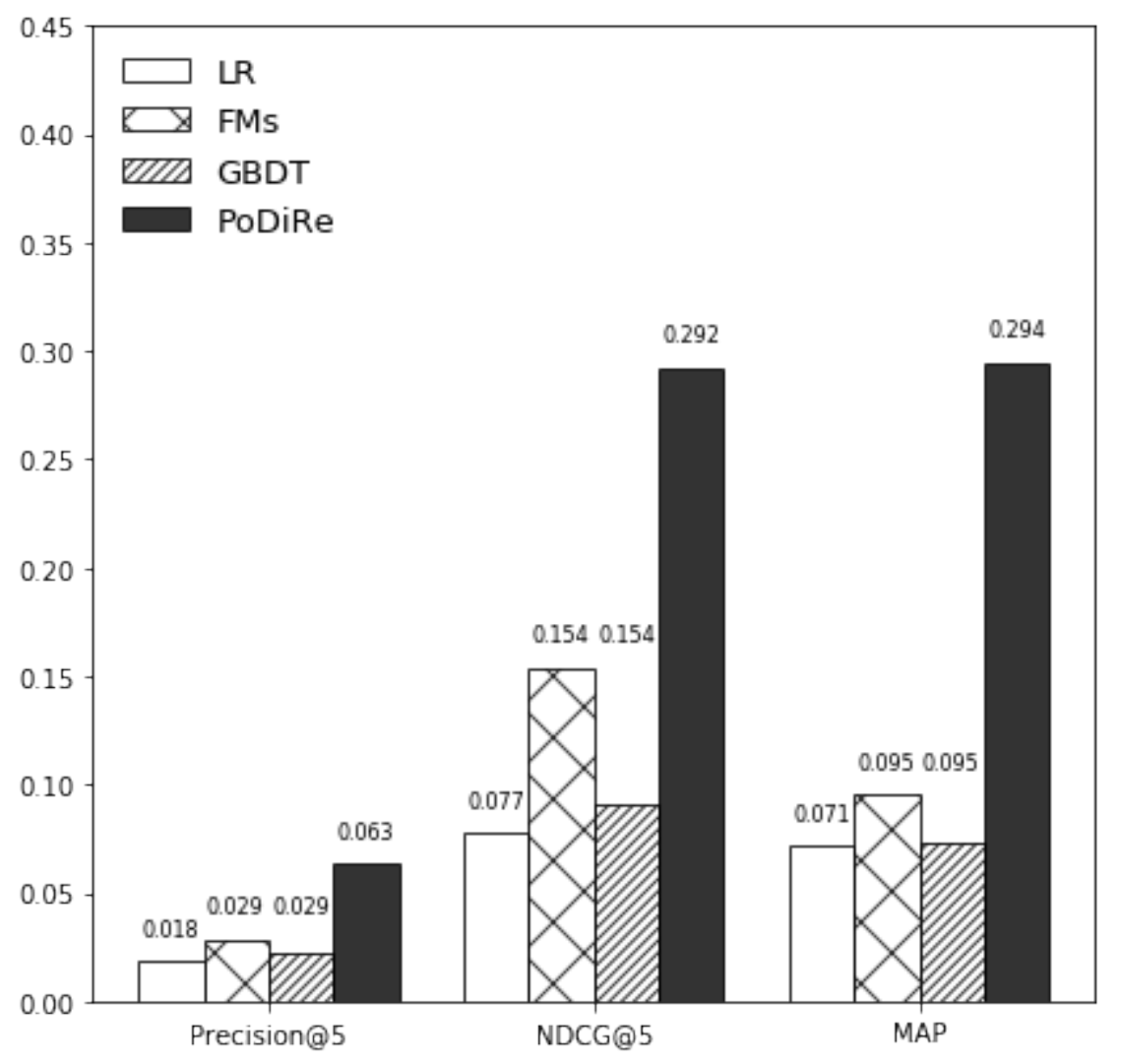}
\label{fig:comparison with single-task supervised learning - install}}
\hfill
\subfigure[Play]{
\includegraphics[width=0.31\textwidth]{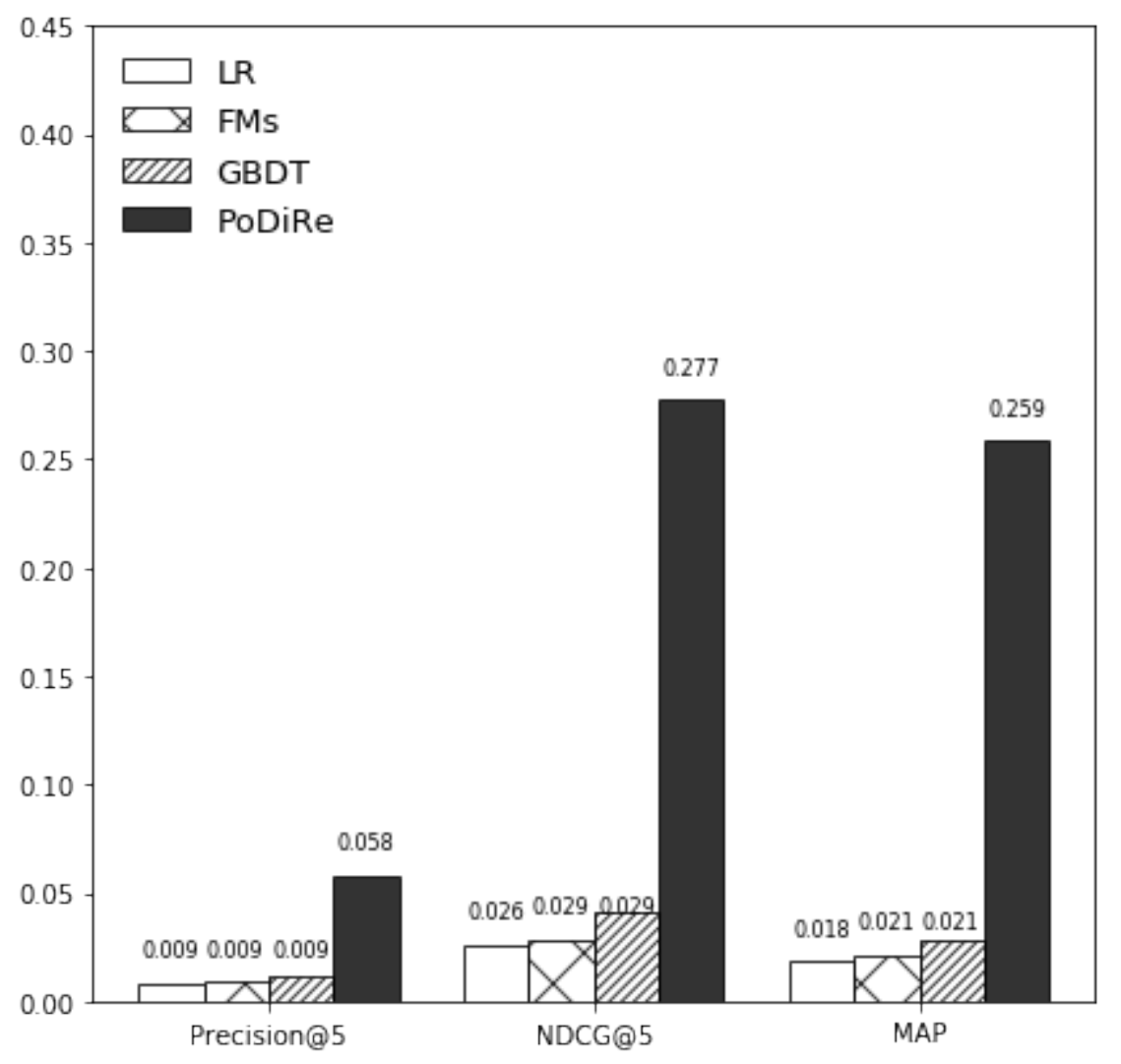}
\label{fig:comparison with single-task supervised learning - play}}
\vspace{-3mm}
\caption{Performance comparison between {\it PoDiRe} and major competitors based on single-task supervised learning over multiple recommendation tasks with $K = 5$}
\label{fig:comparison with single-task supervised learning}
\vspace{-2mm}
\end{figure*}

To evaluate the effectiveness of {\it PoDiRe}, we compare it with nine  competitors over three evaluation metrics ($Precision@K$, $NDCG@K$, and $MAP$). Fig.~\ref{fig:comparison with single-task supervised learning} shows the performance of {\it PoDiRe} against LR, FMs and GBDT when $K=5$. Fig.~\ref{fig:comparison with single-task supervised learning - click}, Fig.~\ref{fig:comparison with single-task supervised learning - install}, and Fig.~\ref{fig:comparison with single-task supervised learning - play}  correspond to the performance on three types of feedback click, install, and play, respectively. It is interesting to note that, simpler methods such as LR and FM perform better than GBDT over the click optimization task and install optimization task but underperform GBDT in the play optimization task. This happens likely due to that the difficulty decreases from the play optimization task to the install optimization task to the click optimization task as the data imbalance and label sparsity are worse from the latter to the former. GBDT is based on a nonlinear classifier and thus can capture the nonlinear relationship between the output and features, which is more suitable to handle a more complex and difficult task. As a comparison, LR and FMs assume the output linearly depends on the first-order or second-order interactions of the features, the underlying assumptions of which are more likely to hold in simpler tasks. Another observation for Fig.~\ref{fig:comparison with single-task supervised learning} is that {\it PoDiRe} substantially outperforms all the three baseline methods over all the recommendation tasks and all evaluation metrics. Its performance is also relatively stable among tasks of different levels of difficulties compared to other competitors. The reason is that all three baseline methods are based on supervised learning and thus unable to plan the recommendations in a way that considers the long-term reward. They also follow the single-task learning framework and thus fail to take advantage of the knowledge from other tasks to improve its performance of each task. On the contrary, {\it PoDiRe} makes recommendations by optimizing the rewards in the longer horizon and encourages knowledge-sharing through jointly learning multiple recommendation tasks, and is trained with extra data generated by well-trained teachers.

\begin{figure*}[htbp]
\centering
\subfigure[Click]{
\includegraphics[width=0.31\textwidth]{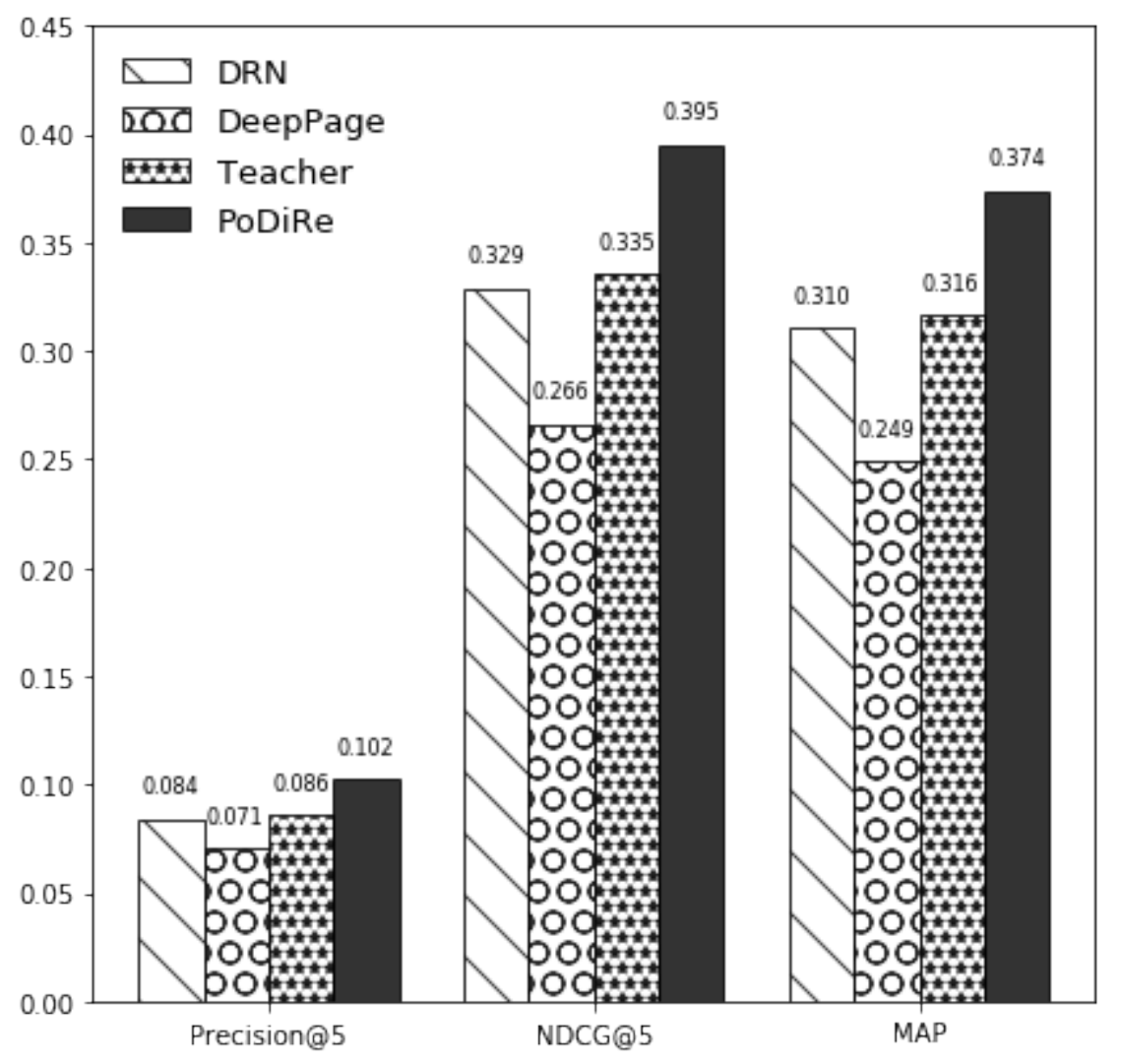}
\label{fig:comparison with single-task reinforcement learning - click}}
\hfill
\subfigure[Install]{
\includegraphics[width=0.31\textwidth]{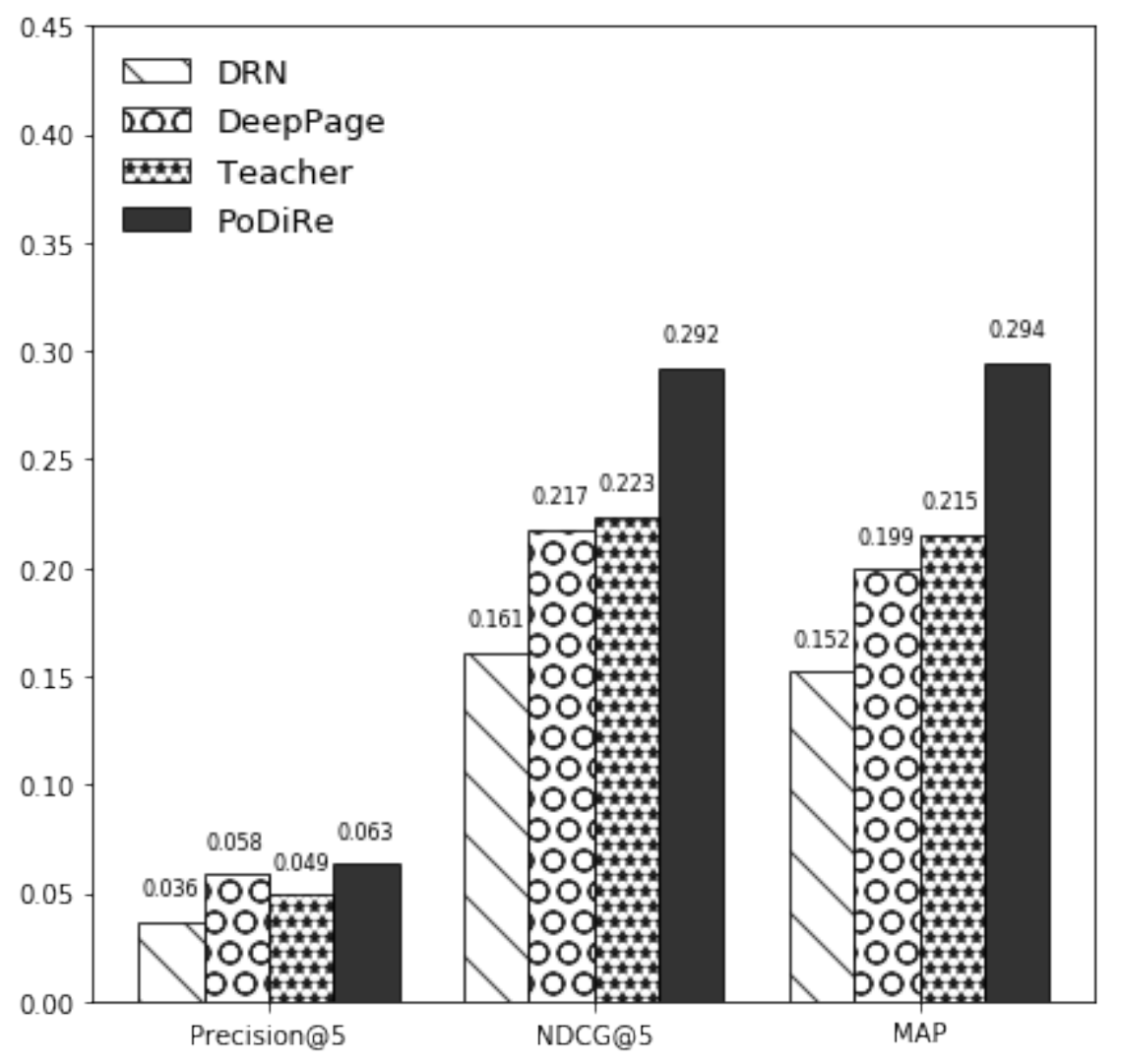}
\label{fig:comparison with single-task reinforcement learning - install}}
\hfill
\subfigure[Play]{
\includegraphics[width=0.31\textwidth]{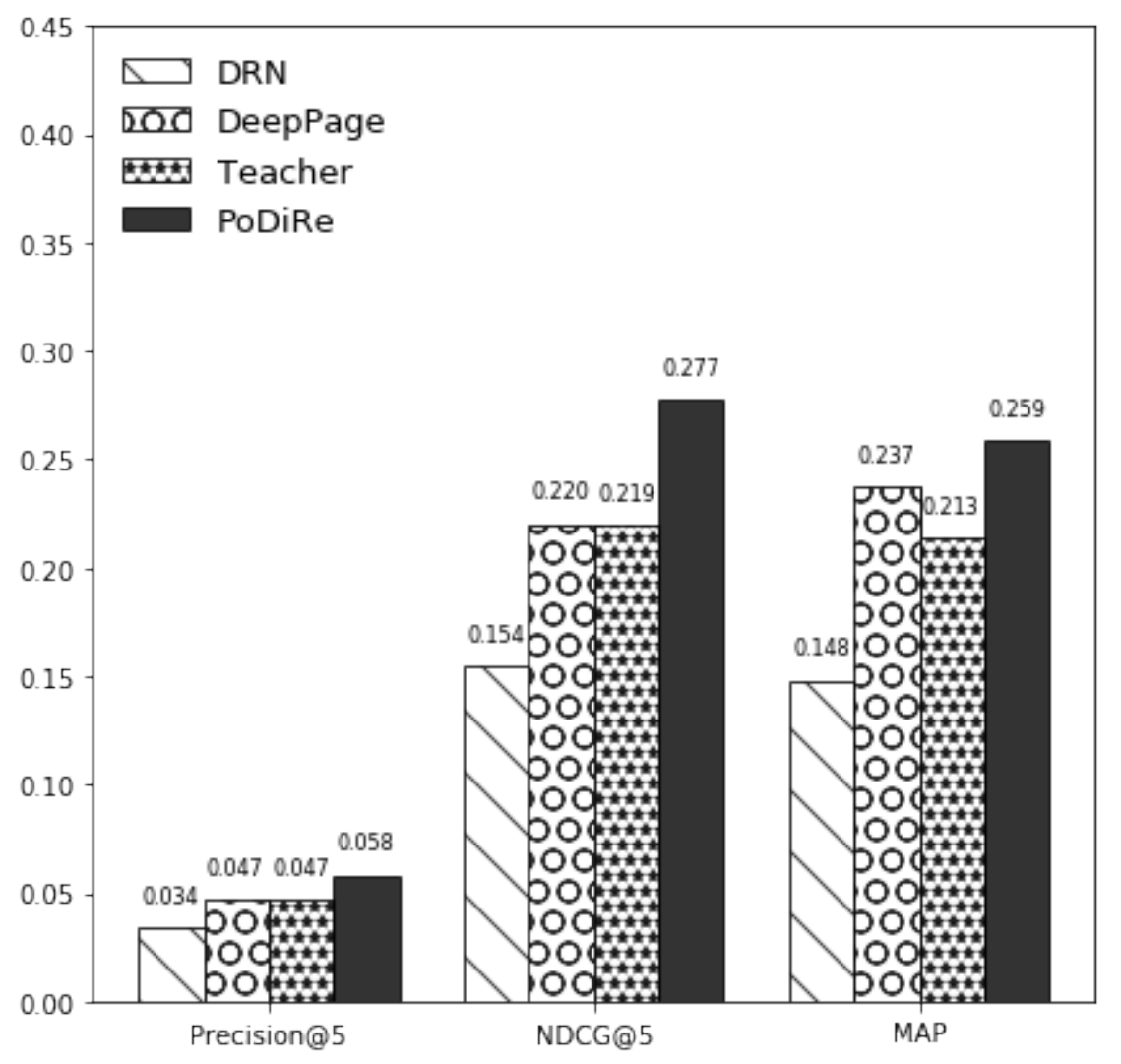}
\label{fig:comparison with single-task reinforcement learning - play}}
\vspace{-3mm}
\caption{Performance comparison between {\it PoDiRe} and major competitors based on single-task reinforcement learning over multiple recommendation tasks with $K = 5$}
\label{fig:comparison with single-task reinforcement learning}
\vspace{-2mm}
\end{figure*}

Fig.~\ref{fig:comparison with single-task reinforcement learning} illustrates the performance of {\it PoDiRe} against DRN, DeepPage, and the Teacher Network over the same collection of recommendation tasks and evaluation metrics as in Fig.~\ref{fig:comparison with single-task supervised learning}. These three baseline methods are based on single-task deep reinforcement learning. Their primary difference falls into the way of state representation. Compared to DRN that handcrafts the features in states, DeepPage learns the state representation by inputting the most recent $T$ items into RNNs. Consequently, DRN outperforms DeepPage in an easy task (click optimization) but underperforms DeepPage in more difficult tasks (install and play optimization). Along with the state representation by DRN that captures the short-term interest of a user, the Teacher Network proposed by this paper introduces into a state a relevant and relatively slowly-changing part that captures the long-term interest of the user. As a result, the performance of the Teacher Network is similar or slightly better than that of DRN over almost all tasks and evaluation metrics. It is worth emphasizing that {\it PoDiRe} has obvious advantage over most tasks and evaluation metrics compared to the three baseline methods, and this advantage becomes more obvious in more difficult tasks as shown in Fig.~\ref{fig:comparison with single-task reinforcement learning - install} and Fig.~\ref{fig:comparison with single-task reinforcement learning - play}. This is likely because {\it PoDiRe} benefits from the improved state representation as well as knowledge-sharing in the multi-task learning process. Another interesting observation is that compared to the supervised learning baselines presented in Fig.~\ref{fig:comparison with single-task supervised learning}, the three baselines in Fig.~\ref{fig:comparison with single-task reinforcement learning} demonstrate more stable performance over different tasks with the same evaluation metrics. The reason is as aforementioned: compared to SL-based recommenders that optimize short-term rewards, the RL-based recommenders plan their recommendations to optimize a long-term goal that may overcome the uncertainty when task difficulty varies.

\begin{figure*}[htbp]
\centering
\subfigure[Click]{
\includegraphics[width=0.31\textwidth]{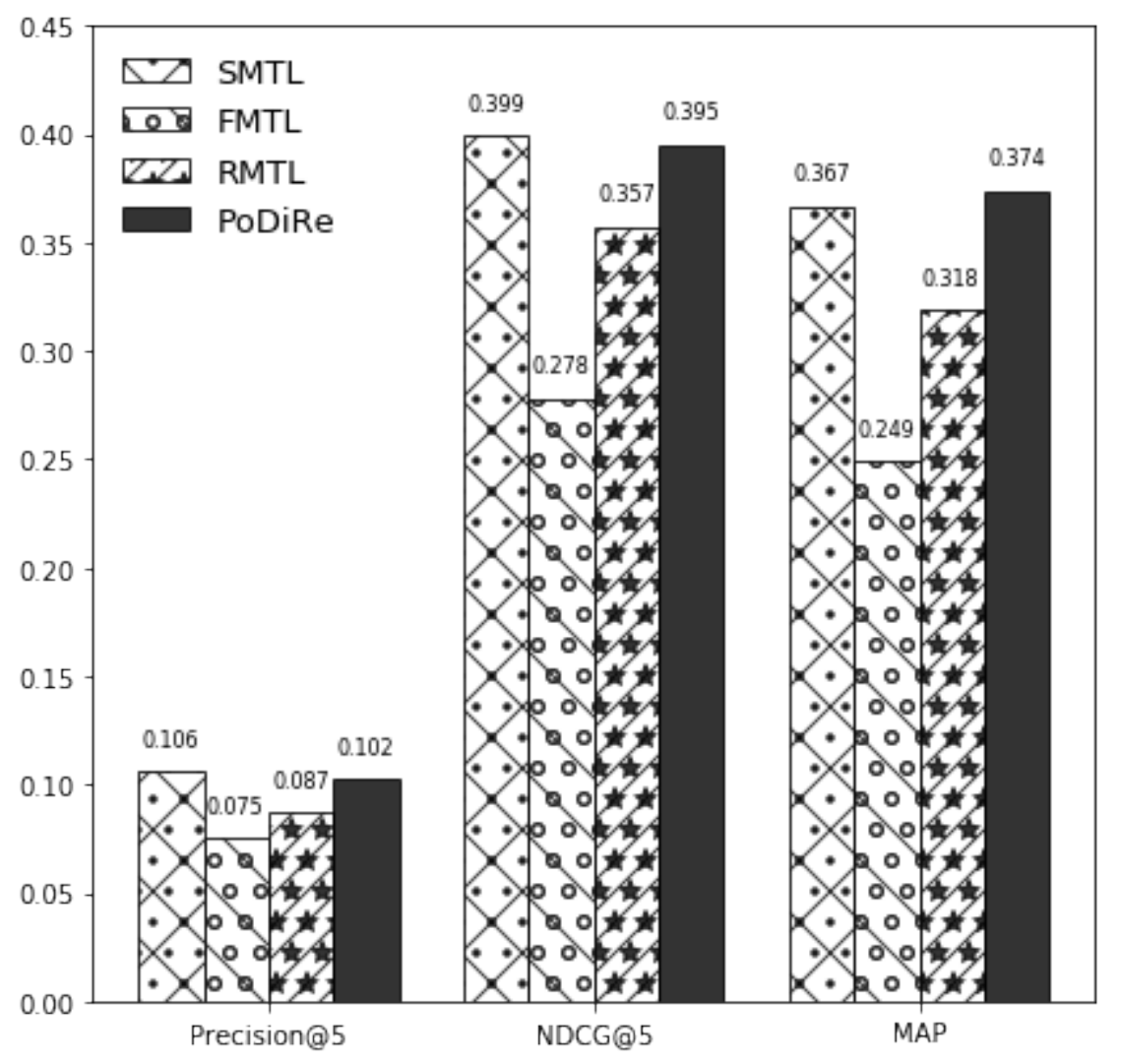}
\label{fig:comparison with multi-task supervised learning - click}}
\hfill
\subfigure[Install]{
\includegraphics[width=0.31\textwidth]{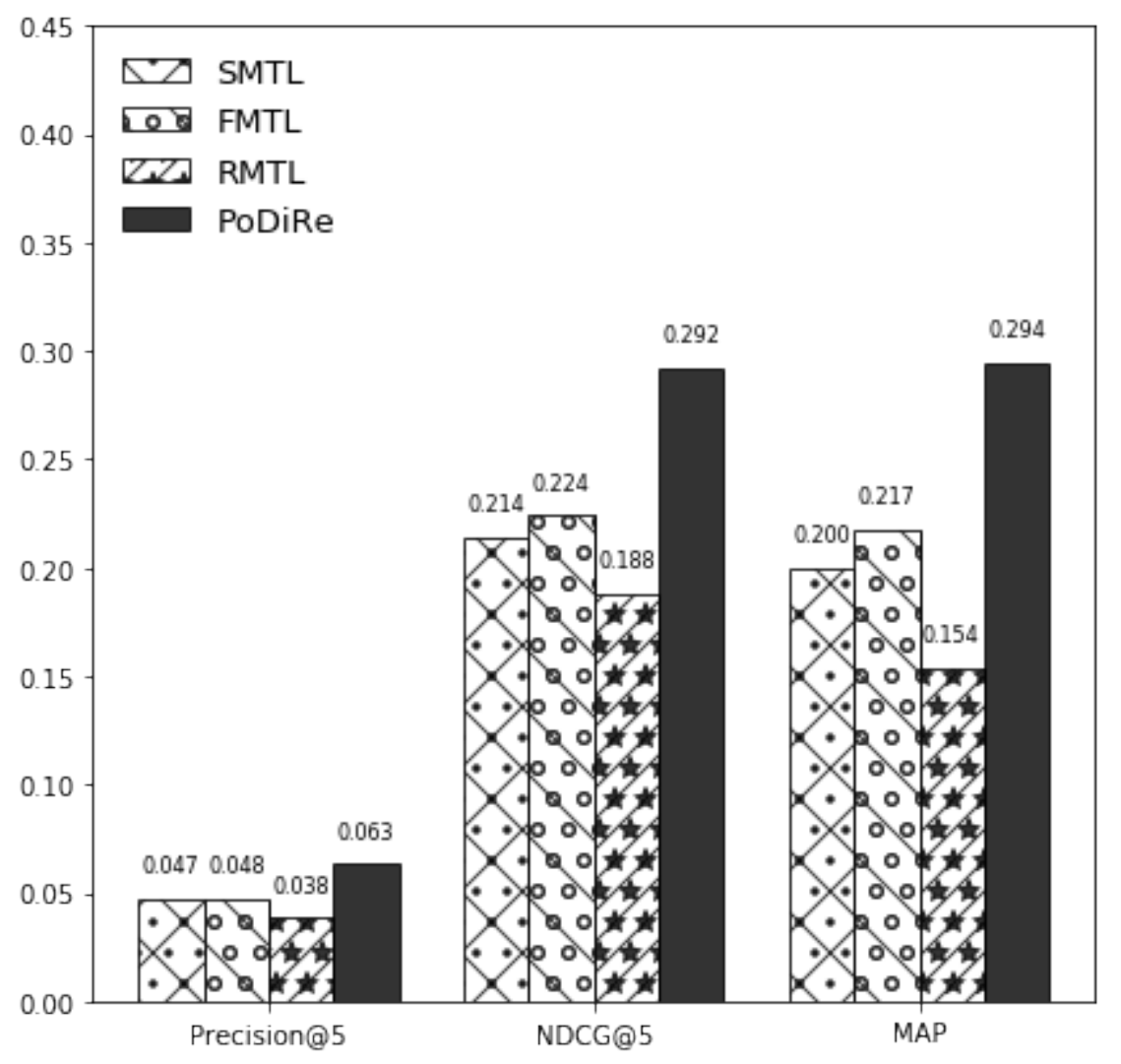}
\label{fig:comparison with multi-task supervised learning - install}}
\hfill
\subfigure[Play]{
\includegraphics[width=0.31\textwidth]{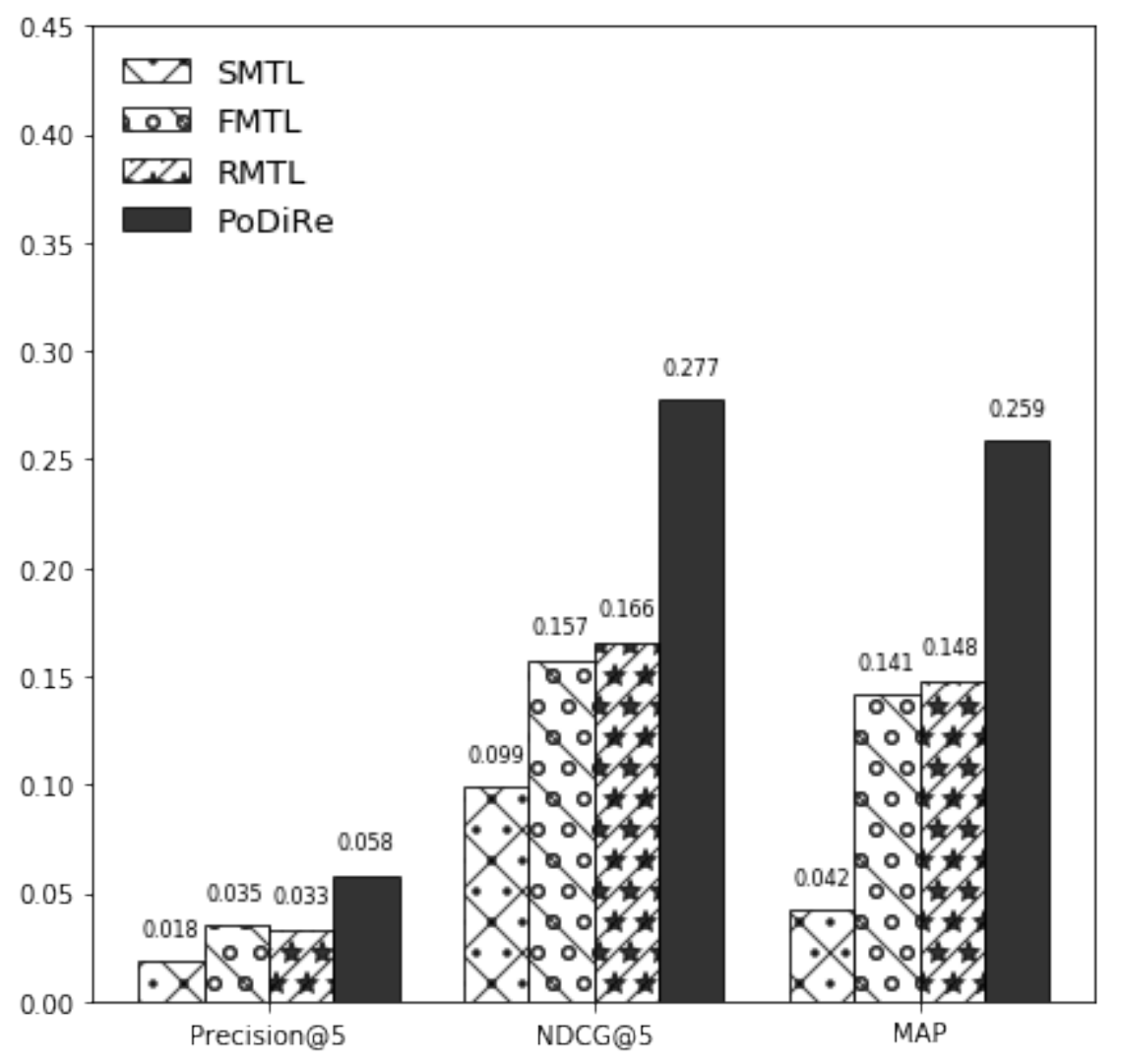}
\label{fig:comparison with multi-task supervised learning - play}}
\vspace{-3mm}
\caption{Performance comparison between {\it PoDiRe} and major competitors based on multi-task supervised learning over multiple recommendation tasks with $K = 5$}
\label{fig:comparison with multi-task supervised learning}
\vspace{-2mm}
\end{figure*}

Fig.~\ref{fig:comparison with multi-task supervised learning} shows the performance of {\it PoDiRe} against SMTL, FMTL, and RMTL over the same collections tasks and evaluation metrics as in above two comparisons. The differences between these three baselines are their ways to capture task relatedness. SMTL assumes that related tasks share a common set of features and thus intends to rule out the unrelated features across tasks by shrinking the entire rows of the matrix to zero. This assumption is more likely to hold in simple tasks. Unfortunately, it can be easily violated in complex applications. To overcome its limitation, FMTL introduces another term along with the existing one in SMTL. The new term allows for different tasks to learn task-specific features. In other words, FMTL offers more freedom in knowledge transfer from easy tasks to difficult tasks. As illustrated in Fig.~\ref{fig:comparison with multi-task supervised learning}, FMTL demonstrates obvious advantages than SMTL in the difficult tasks, as shown in Fig.~\ref{fig:comparison with multi-task supervised learning - install} and Fig.~\ref{fig:comparison with multi-task supervised learning - play}. Different from SMTL and FMTL, RMTL believes that the weights of all tasks are close to each other. This belief is established by minimizing the distance between the weights vectors of different tasks and a shared weights vector. Compared to the other two baselines, RMTL seems to favor more difficult tasks than easy ones in knowledge transfer. As presented by Fig.~\ref{fig:comparison with multi-task supervised learning - play}, it performs better in the play-optimization task than SMTL and FMTL. One crucial observation is that these three baselines outperform LR, FMs and GBDT, baselines based on single-task supervised learning, over almost all tasks and evaluation metrics, if one compares Fig.~\ref{fig:comparison with multi-task supervised learning} with Fig.~\ref{fig:comparison with single-task supervised learning}. This advantage is largely because of the knowledge-sharing between different tasks. It is also worth emphasizing that {\it PoDiRe} achieves outstanding performance over all tasks and almost all evaluation metrics compared to the three baselines. The reason is that although both {\it PoDiRe} and the baselines follow MTL framework, {\it PoDiRe} aims to maximize long-term rewards in making recommendations and takes into account the impacts of current recommendation to future rewards. In all the above comparisons between {\it PoDiRe} and nine baselines, the parameter of the evaluation metrics is set as $K=5$. We also conduct comparisons with $K=10$ and include them in Appendix~\ref{appendix: additional simulations}, where similar trends of comparisons can be observed.

\subsection{Efficiency of {\it PoDiRe}}\label{subsec: efficiency}

\begin{table}[htbp]
\centering
\caption{Comparison between \textit{PoDiRe} and major competitors in model size and response time.}
\label{tab:response_time}
\begin{tabular}{|p{0.28\columnwidth}|p{0.19\columnwidth}|p{0.17\columnwidth}|}
\hline
& Model Size (\# of parameters) & Response Time (ms)   \\ \hline
{\it PoDiRe}        & {\textbf{5,715}}          & {\textbf{10.96}} \\ \hline
Teacher       & 8,430                         & 34.08                \\ \hline
DeepPage       & 8,100                         & 29.88                \\ \hline
DRN       & 8,755                         & 33.27                \\ \hline
RMTL/SMTL/FMTL & 15,670                        & 20.08                \\ \hline
\end{tabular}
\end{table}



To evaluate the efficiency of {\it PoDiRe}, we compare the model size and average response time between {\it PoDiRe} and representatives baseline methods in Table~\ref{tab:response_time}. In the table, model size refers to the total number of to-be-learned parameters after the optimization of hyperparameters through grid search. The response time is computed as the total time in milliseconds to generate recommendations to users divided by the total number of users. As displayed in Table~\ref{tab:response_time}, {\it PoDiRe} outperforms all the representative baseline methods in terms of the model size and response time. For example, {\it PoDiRe} is only $2/3$ size of a single teacher network, and its average response time is reduced to $1/3$ of the teacher network. This largely benefits from knowledge sharing in jointly learning multiple tasks. It also benefits from the extra training data generated by the well-trained teachers. Note that in the comparison, we ignore LR, FMs, and GBDT because they have non-DNN structures, and their effectiveness is much less satisfactory than {\it PoDiRe}. 

\subsection{Effects of the Long-Term Interest Vector}
\label{subsec:long-term interest}

\begin{table}[htbp]
\caption{Effects of long-term interest vector on the performance of the Teacher}
\label{tab:long-term interest}
\begin{tabular}{|l|c|c|c|c|c|c|c|c|c|}
\hline
\begin{tabular}[c]{@{}l@{}}Evaluation Metric\end{tabular}      & \multicolumn{3}{c|}{Precision@5}                 & \multicolumn{3}{c|}{NDCG@5}                      & \multicolumn{3}{c|}{MAP}                         \\ \hline
Task                                                              & Click          & Install        & Play           & Click          & Install        & Play           & Click          & Install        & Play           \\ \hline
Short-Term                                                         & 0.084          & 0.041          & 0.040          & 0.328          & 0.181          & 0.190          & 0.310          & 0.204          & 0.197          \\ \hline
Long-Term                                                         & 0.079          & 0.045          & 0.043          & 0.295          & 0.220          & 0.211          & 0.292          & 0.209          & 0.211          \\ \hline
\begin{tabular}[c]{@{}l@{}}Short-Term +\\  Long-Term\end{tabular} & \textbf{0.086} & \textbf{0.049} & \textbf{0.047} & \textbf{0.335} & \textbf{0.223} & \textbf{0.219} & \textbf{0.316} & \textbf{0.215} & \textbf{0.213} \\ \hline
\end{tabular}
\end{table}

To examine the attribution of the long-term interest vector in the state representation to the outstanding performance of {\it PoDiRe}, we conduct an ablative study in Table~\ref{tab:long-term interest}. Since the teachers and student share the same state representation and the performance of the student is often in proportional to the performance of its teachers, the study only compares the performance of Teacher when different parts of the represented state are utilized in learning the recommendation policy. In Table~\ref{tab:long-term interest}, ``Short-Term'' means that the long-term interest vector $\bm{u}_{\ell}$ is not included in the state representation $\bm{s}_{t}$, i.e., $\bm{s}_{t}:= \text{concat}(\bm{u}_{s}, \bm{u}_{c})$. Similarly, ``Long-Term'' indicates that $\bm{s}_{t}:= \text{concat}(\bm{u}_{\ell}, \bm{u}_{c})$ and ``Short-Term + Long-Term'' implies that both the short-term interest vector and the long-term interest vector are used in representing the state, i.e., $\bm{s}_{t}:= \text{concat}(\bm{u}_{s},\bm{u}_{\ell}, \bm{u}_{c})$. As shown in the table, the model that uses both long-term interest vector and short-term interest vector in training outperforms those using one of them. This validates that the introduction of long-term interest vector into the state representation in Section~\ref{subsec:state_representation} plays a positive role in improving the overall performance.

\subsection{Effects of Different Teachers on the Performance of Student}\label{subsec: different teachers}

\begin{table}[htbp]
\caption{Effects of different teachers on the performance of the Student}
\label{tab: effects of teachers to student}
\begin{tabular}{|l|c|c|c|c|c|c|c|c|c|}
\hline
Evaluation Metric   & \multicolumn{3}{c|}{Precision@5}                 & \multicolumn{3}{c|}{NDCG@5}                      & \multicolumn{3}{c|}{MAP}                         \\ \hline
Task                & Click          & Install        & Play           & Click          & Install        & Play           & Click          & Install        & Play           \\ \hline
Trained by DRN      & 0.101          & 0.053          & 0.044          & 0.389          & 0.234          & 0.203          & 0.371          & 0.225          & 0.175          \\ \hline
Trained by DeepPage & 0.095          & \textbf{0.069} & 0.058          & 0.328          & 0.290          & \textbf{0.287} & 0.323          & 0.275          & \textbf{0.270} \\ \hline
Trained by the prop-  & \textbf{0.102} & 0.063          & \textbf{0.058} & \textbf{0.395} & \textbf{0.292} & 0.277          & \textbf{0.374} & \textbf{0.294} & 0.259 \\
osed Teacher Network  &  &           &  &  &  &           &  &  &       \\ \hline
\end{tabular}
\end{table}

To demonstrate the application of the proposed Teacher Network in training {\it PoDiRe}, we compare the performance of the Student Network trained by different teachers in Table~\ref{tab: effects of teachers to student}. The teachers compared with were also illustrated in Fig.~\ref{fig:comparison with single-task reinforcement learning}. Here a student is ``trained'' by a teacher means that the training data for the student is generated by the teacher model. As illustrated by Table~\ref{tab: effects of teachers to student}, compared to students trained by other teachers, the student that is trained by the proposed Teacher Network achieves the best performance over almost all recommendation tasks and all evaluation metrics. 

\subsection{Effects of Policy Distillation}
\label{subsec:effects of distillation}

\begin{figure*}[htpb]
\centering
\subfigure[$K = 5$]{
\includegraphics[width=0.46\textwidth]{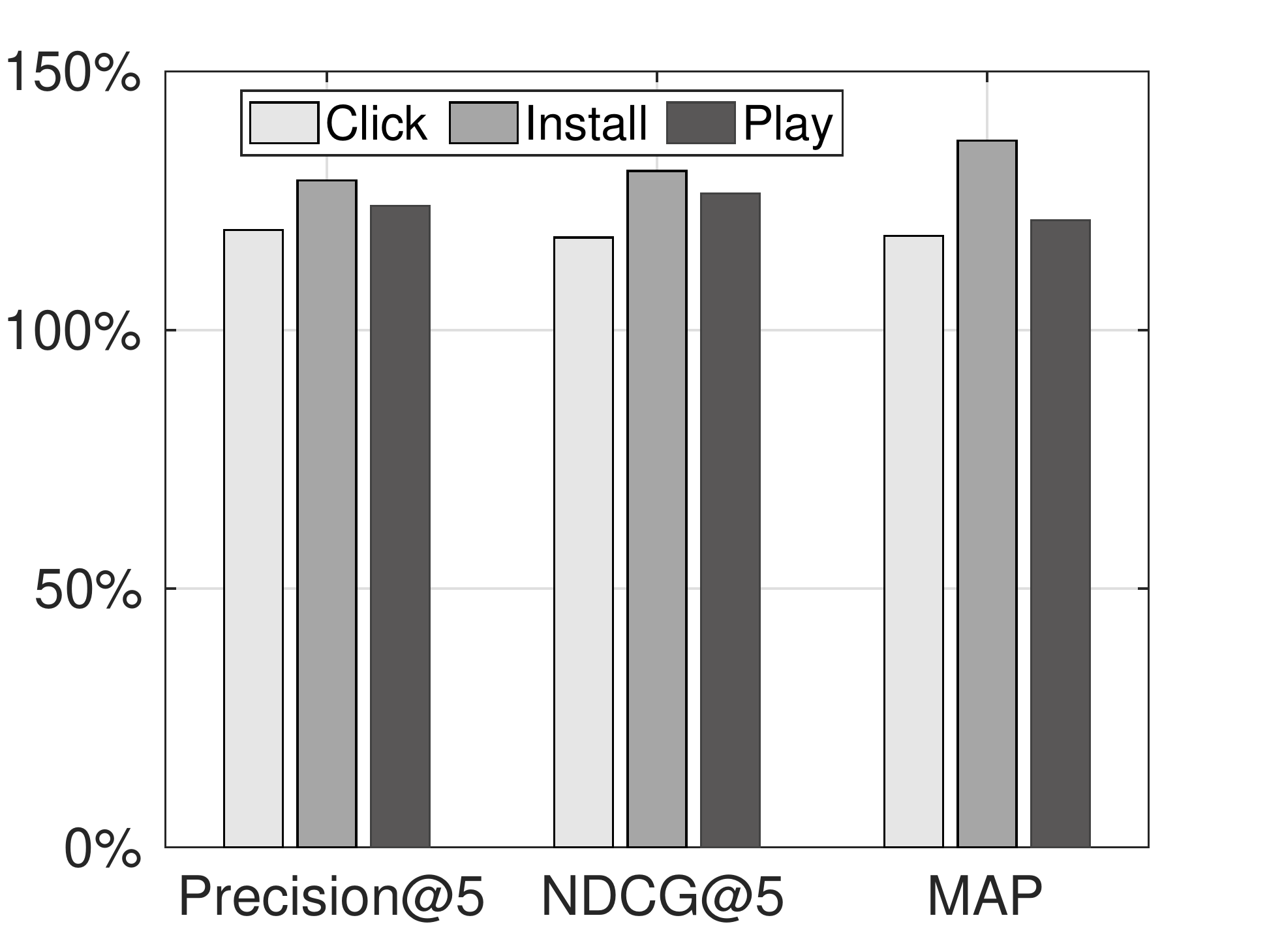}
\label{figure:student_vs_teacher5}}
\hfill
\subfigure[$K = 10$]{
\includegraphics[width=0.46\textwidth]{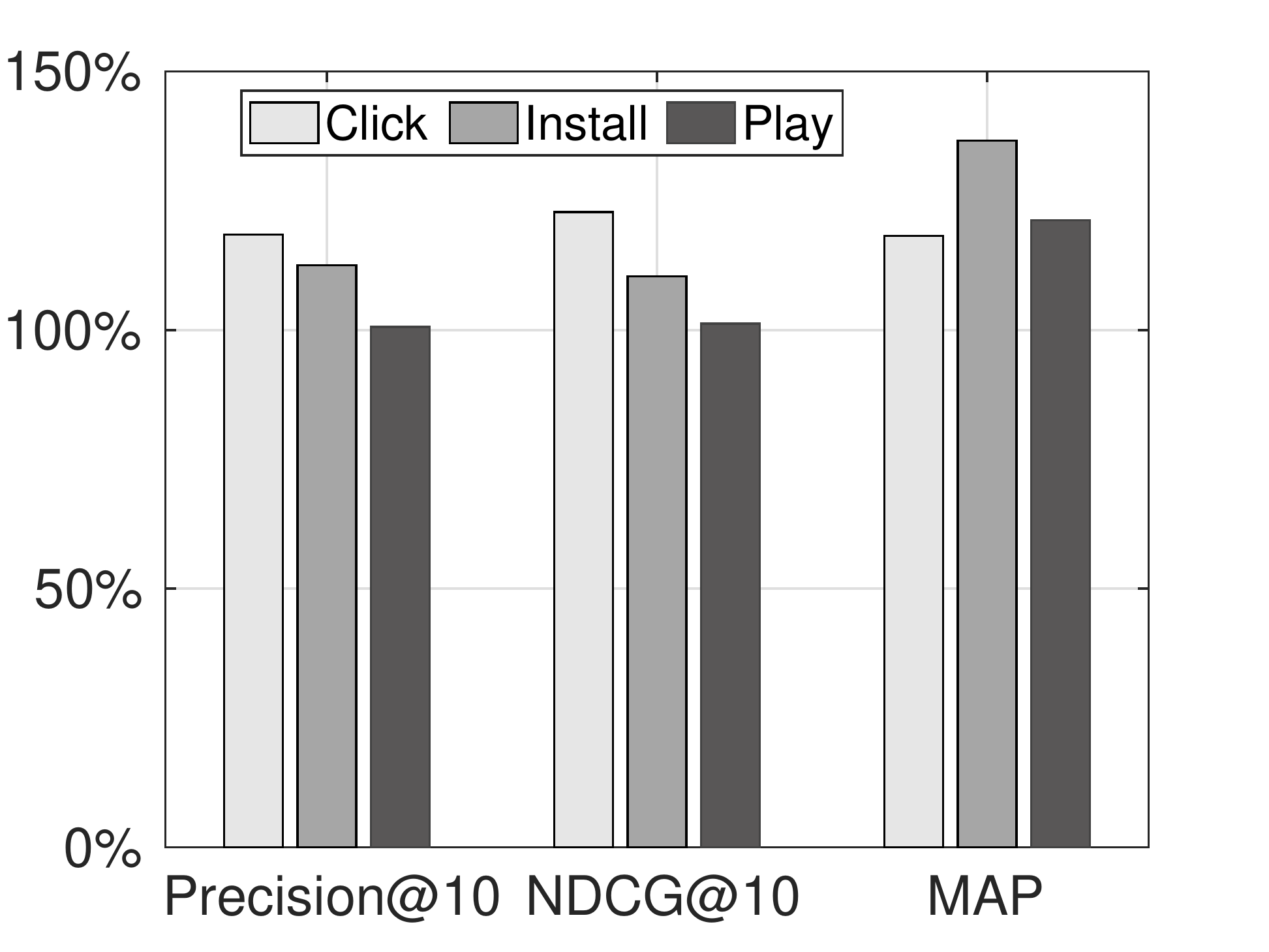}
\label{figure:student_vs_teacher10}}
\hfill
\vspace{-3mm}
\caption{Performance comparison between student and teachers when $K = 5$ and $K=10$. The performance of the teacher on the corresponding task is counted as 100\%}
\label{fig:student_to_teacher}
\vspace{-2mm}
\end{figure*}

To evaluate the effects of policy distillation, we compare the performance of the student and teachers on the three recommendation tasks. The results are shown in Fig.~\ref{fig:student_to_teacher}, where the student's performance is given as a percentage of teachers' corresponding performance. We can observe that student outperforms teachers almost over all tasks and all evaluation metrics. The performance improvement of the student model well justifies the success of our designed policy distillation and multi-task learning in {\it PoDiRe}.

\subsection{Hyper-Parameter Settings}
\label{subsec:parameter}
For the reason of reproducibility, we give the detailed parameter settings used in training as follows.
\begin{itemize}
    \item Game feature dimension $N_{a}$: 49
    \item Temperature $\tau$: 0.01
    \item Time window $T$: 3
    \item Discount factor $\gamma$: 0.6
    \item Embedded user long-term interest vector dimension $N_{\ell}$: 10
    \item Embedded context feature dimension $N_{c}$: 3
    \item GRU hidden state dimension and embedded user short-term interest vector dimension $N_{h}$: 10
    \item Number of GRU layers in state representation $N_{g}$: 3
    \item Number of feedback types $N_{f}$: 3
    \item Number layers for teacher's Q function $N_q$: 4
    \item Number task-shared layers $N_{s}$: 2
    \item Number task-specific layers $N_{i}$: 2 for any $i$
    \item Weights of loss $\{\lambda_{(i)}\}_{i=1}^{N_f}$: $[0.25, 0.25, 0.5]$
    \item Number of epochs $T_{e}$: 5-10
    \item Batch size in training teacher network $N_b$: 64
    \item Learning rate $\{\eta_i\}_{i=1}^{N_f}$ and $\eta_s$: the initial value is 0.01 and decays by $\eta = \eta_0/ (1 + p/2)$, where $p$ is the number of epochs 
    \item Buffer size $N_r$: 256
    \item Optimizer: Adam method
    \item Number of time steps between target network update $T_{-}$: 20
    \item Number of trees in GBDT: 20
    \item Search range of maximal depth in GBDT: $[3, 5]$
    \item Search range of maximal number of bins to discrete continuous features for splitting in GBDT: $[16, 64]$
    \item Search range of step size in GBDT: $[0.001, 0.1]$

\end{itemize}

\subsection{System Configurations}
\label{subsec:configuration}
The software dependencies and environment used in our experiments are given below:
\begin{itemize}
    \item Python: 3.6
    \item Tensorflow: 0.12.1
    \item Numpy: 1.12.1
    \item Pyspark/Spark: 2.2.0
    \item Pandas: 0.20.1
    \item Scikit-learn: 0.18.1
\end{itemize}
The hardware configuration for our experiments is:
\begin{itemize}
    \item AWS EC2 Instance: x1.16xlarge 
    \item CPU: 64 cores of 2.3 GHz Intel Xeon E7-8880 v3 Processor
    \item Memory: 976 GiB
\end{itemize}

%% file: 6_conclusion.tex
\section{Conclusion}
\label{section:conclusion}
Driven by the business desideratum of considering long-term rewards of multiple recommendation tasks, in this paper, we proposed a novel method {\it PoDiRe}, \underline{po}licy \underline{di}stilled \underline{re}commender, that can solve multiple recommendation tasks simultaneously and maximize the long-term rewards of recommendation. {\it PoDiRe} was developed based on deep reinforcement learning, policy distillation, and a unique state representation method combining users' short-term interest, long-term preference, and rich context information. We evaluated our method using a large-scale dataset collected from experiments over the Samsung Game Launcher platform. The evaluation results using multiple metrics demonstrate better effectiveness and efficiency of our developed method (i.e., {\it PoDiRe}) against several state-of-the-art methods.



%% file: 7_appendix.tex
\section{Appendix}
\subsection{Additional Results for Effectiveness Evaluation of {\it PoDiRe}} \label{appendix: additional simulations}

\begin{figure*}[htbp]
\centering
\subfigure[Click]{
\includegraphics[width=0.31\textwidth]{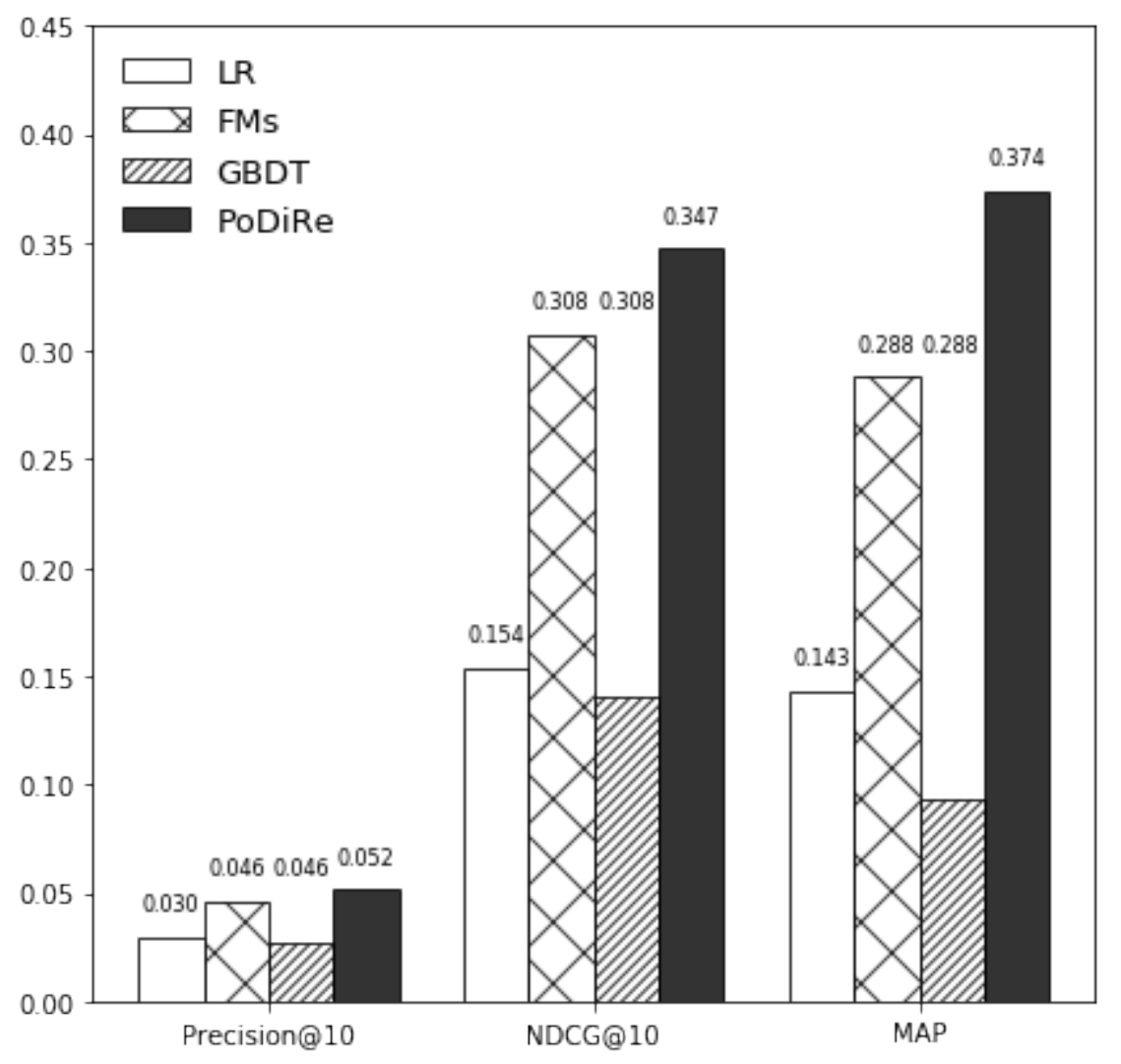}
\label{fig:comparison with single-task supervised learning K=10 - click}}
\hfill
\subfigure[Install]{
\includegraphics[width=0.31\textwidth]{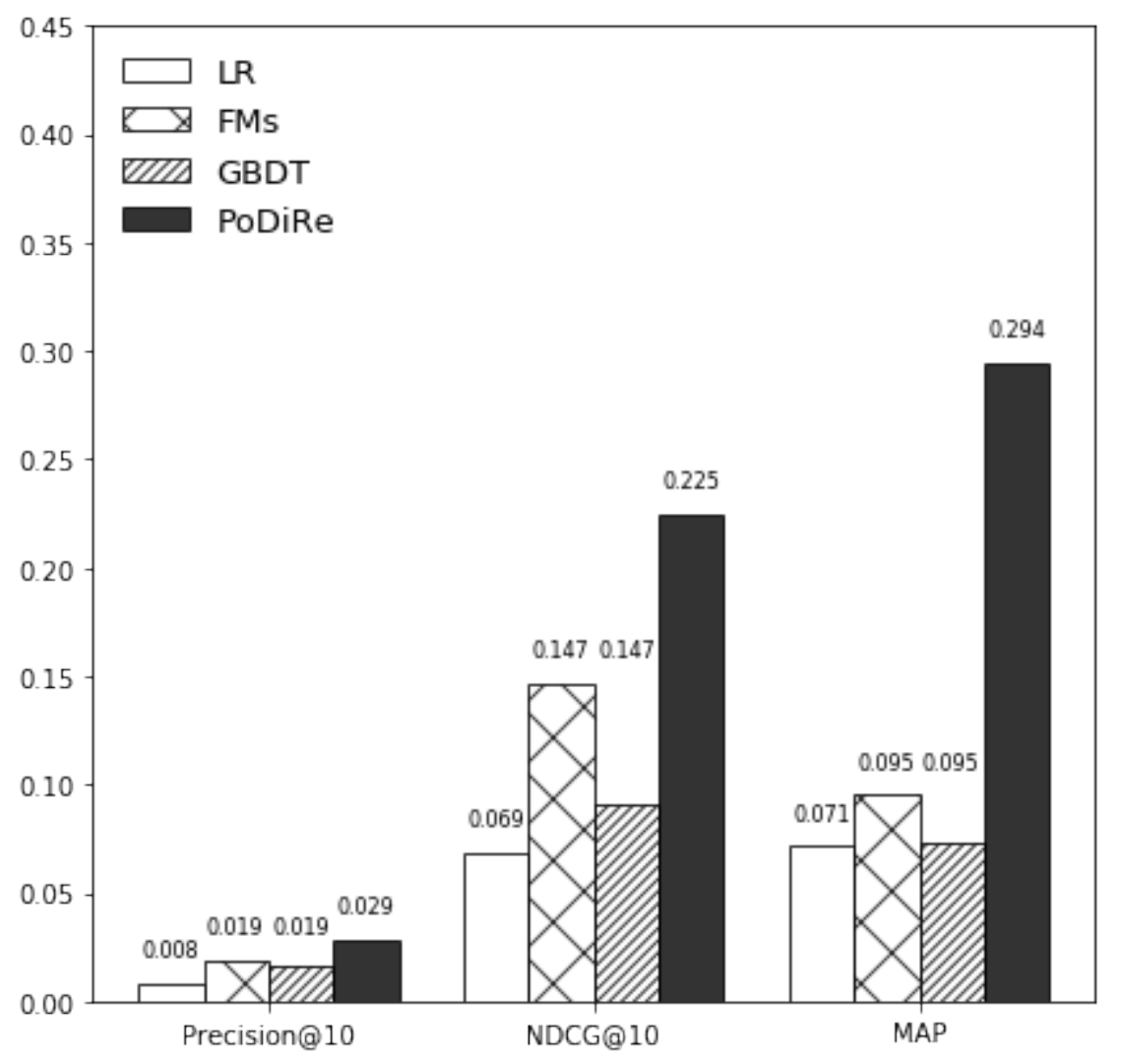}
\label{fig:comparison with single-task supervised learning K=10 - install}}
\hfill
\subfigure[Play]{
\includegraphics[width=0.31\textwidth]{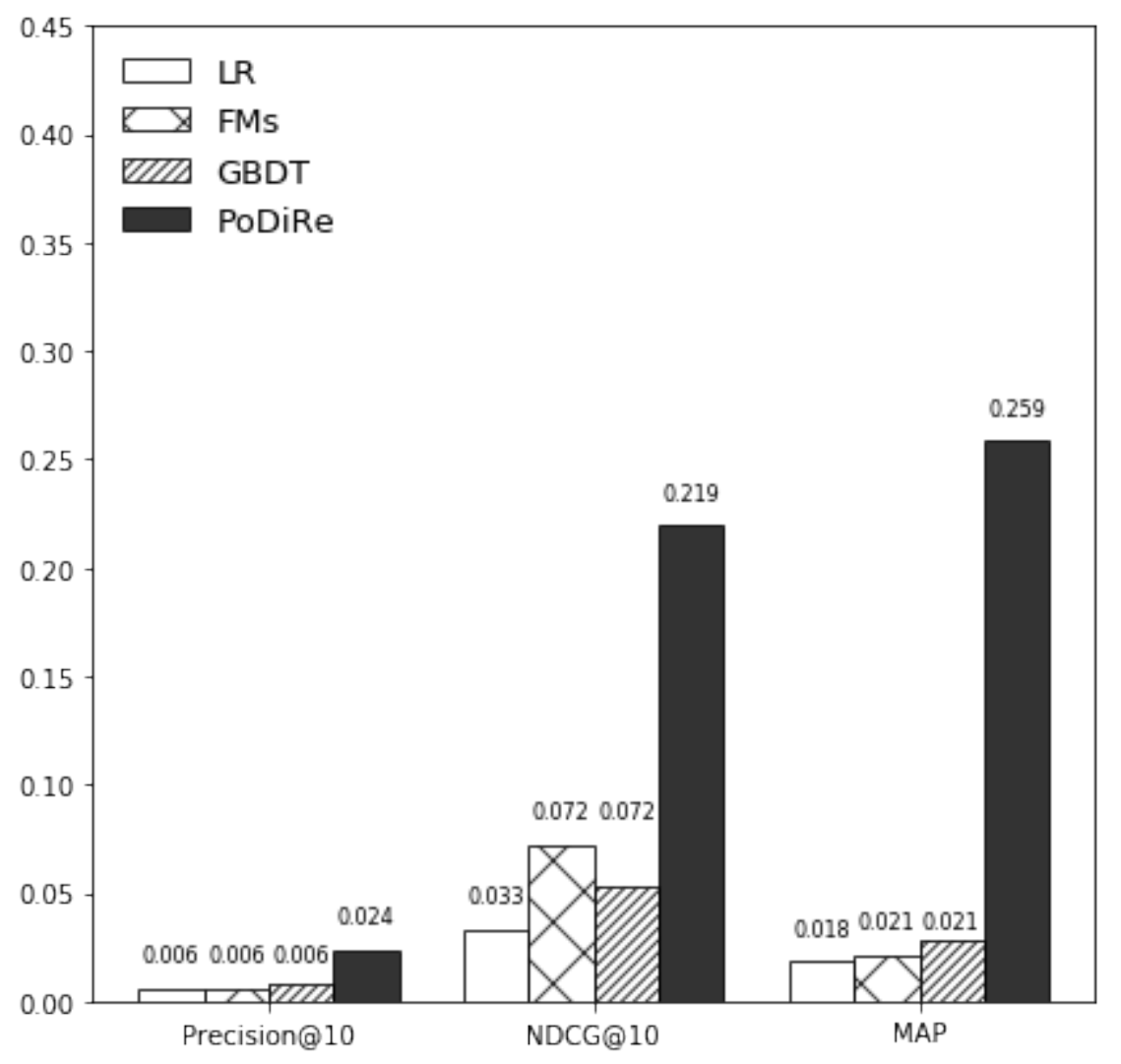}
\label{fig:comparison with single-task supervised learning K=10 - play}}
\vspace{-3mm}
\caption{Performance comparison between {\it PoDiRe} and major competitors based on single-task supervised learning over multiple recommendation tasks with $K = 10$}
\label{fig:comparison with single-task supervised learning K=10}
\vspace{-2mm}
\end{figure*}

\begin{figure*}[htbp]
\centering
\subfigure[Click]{
\includegraphics[width=0.31\textwidth]{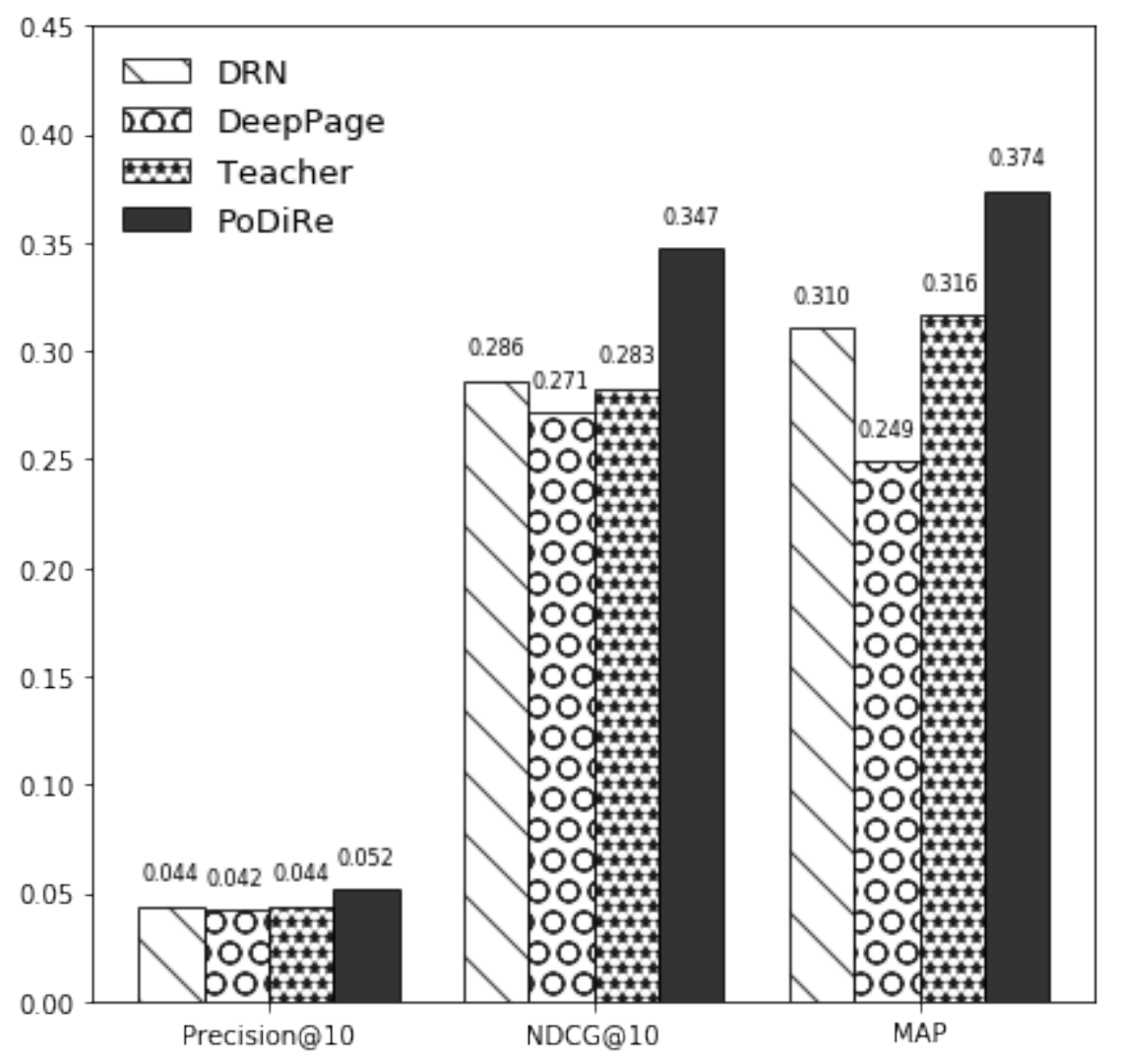}
\label{fig:comparison with single-task reinforcement learning K=10 - click}}
\hfill
\subfigure[Install]{
\includegraphics[width=0.31\textwidth]{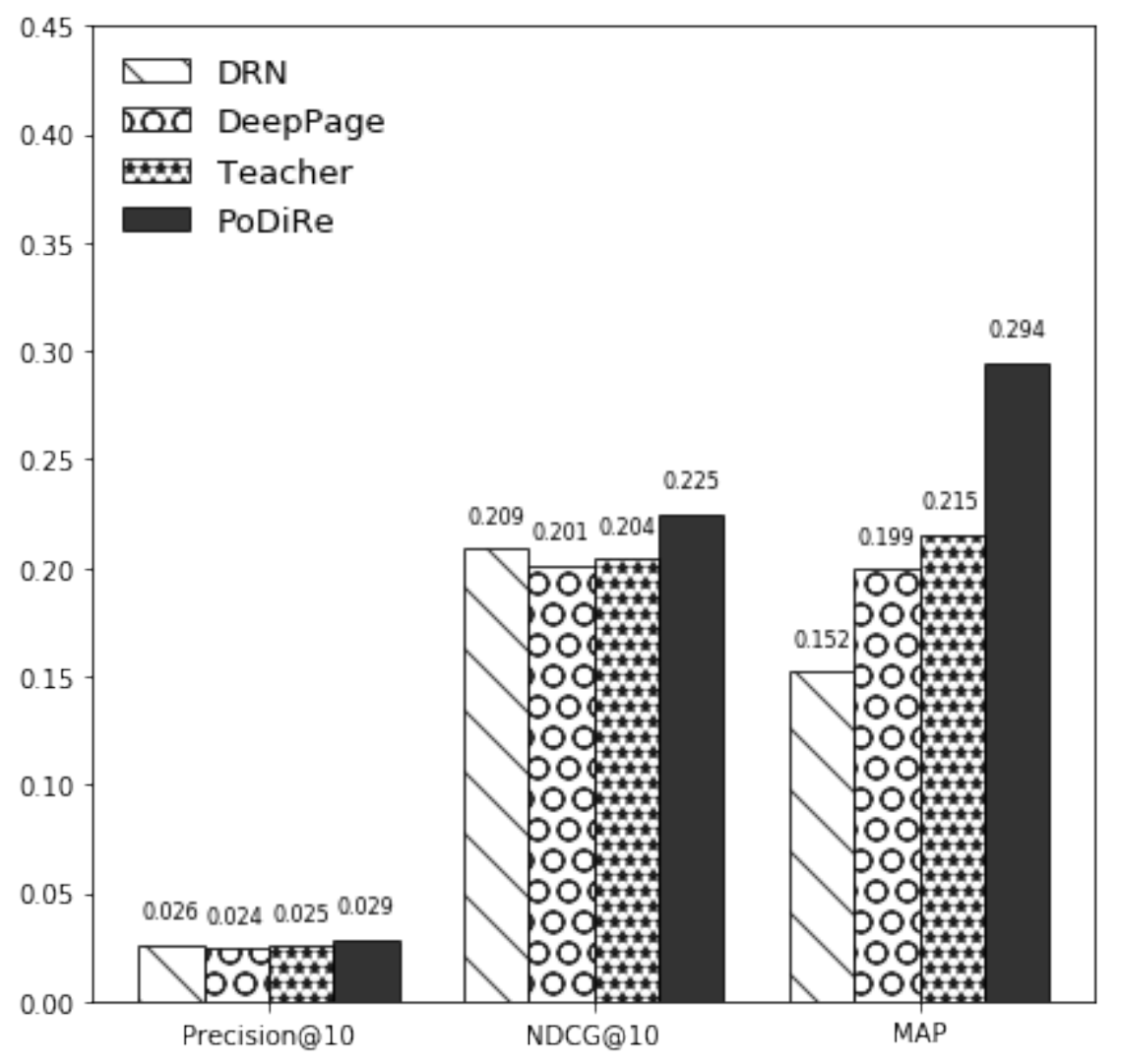}
\label{fig:comparison with single-task reinforcement learning K=10 - intall}}
\hfill
\subfigure[Play]{
\includegraphics[width=0.31\textwidth]{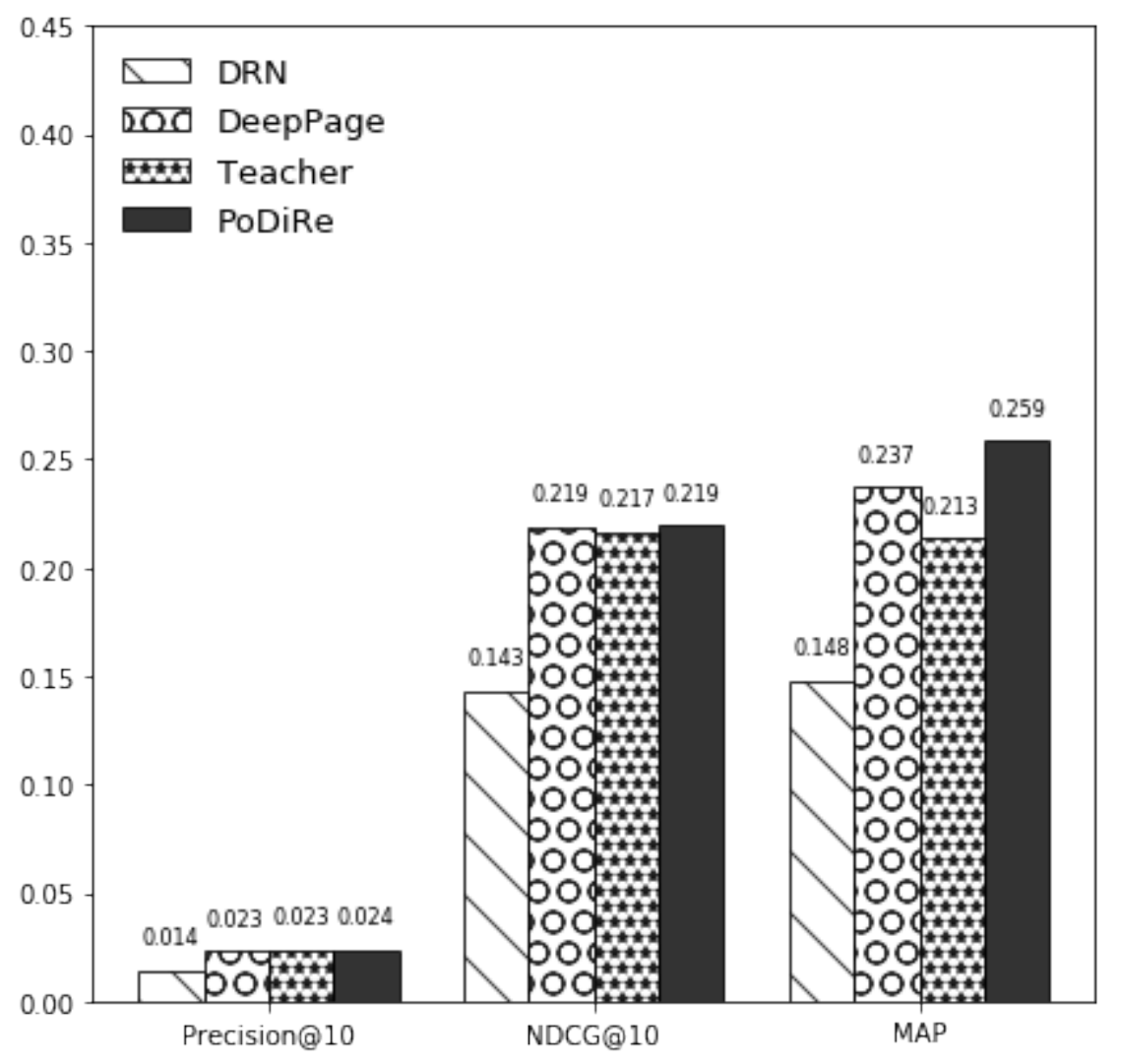}
\label{fig:comparison with single-task reinforcement learning K=10 - play}}
\caption{Performance comparison between {\it PoDiRe} and major competitors based on single-task reinforcement learning over multiple recommendation tasks with $K = 10$}
\label{fig:comparison with single-task reinforcement learning K=10}
\end{figure*}

\begin{figure*}[htbp]
\centering
\subfigure[Click]{
\includegraphics[width=0.31\textwidth]{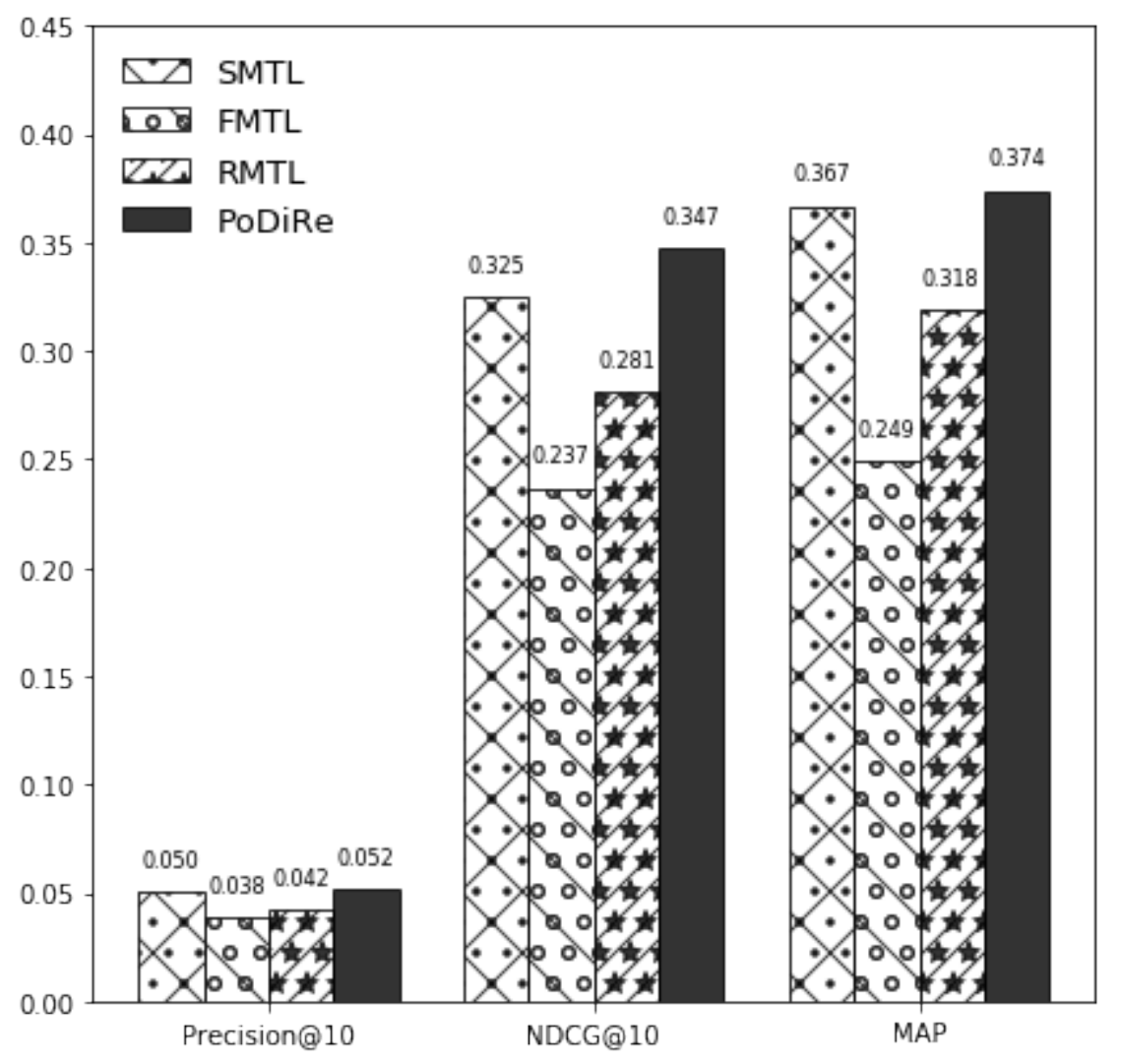}
\label{fig:comparison with multi-task supervised learning K=10 - click}}
\hfill
\subfigure[Install]{
\includegraphics[width=0.31\textwidth]{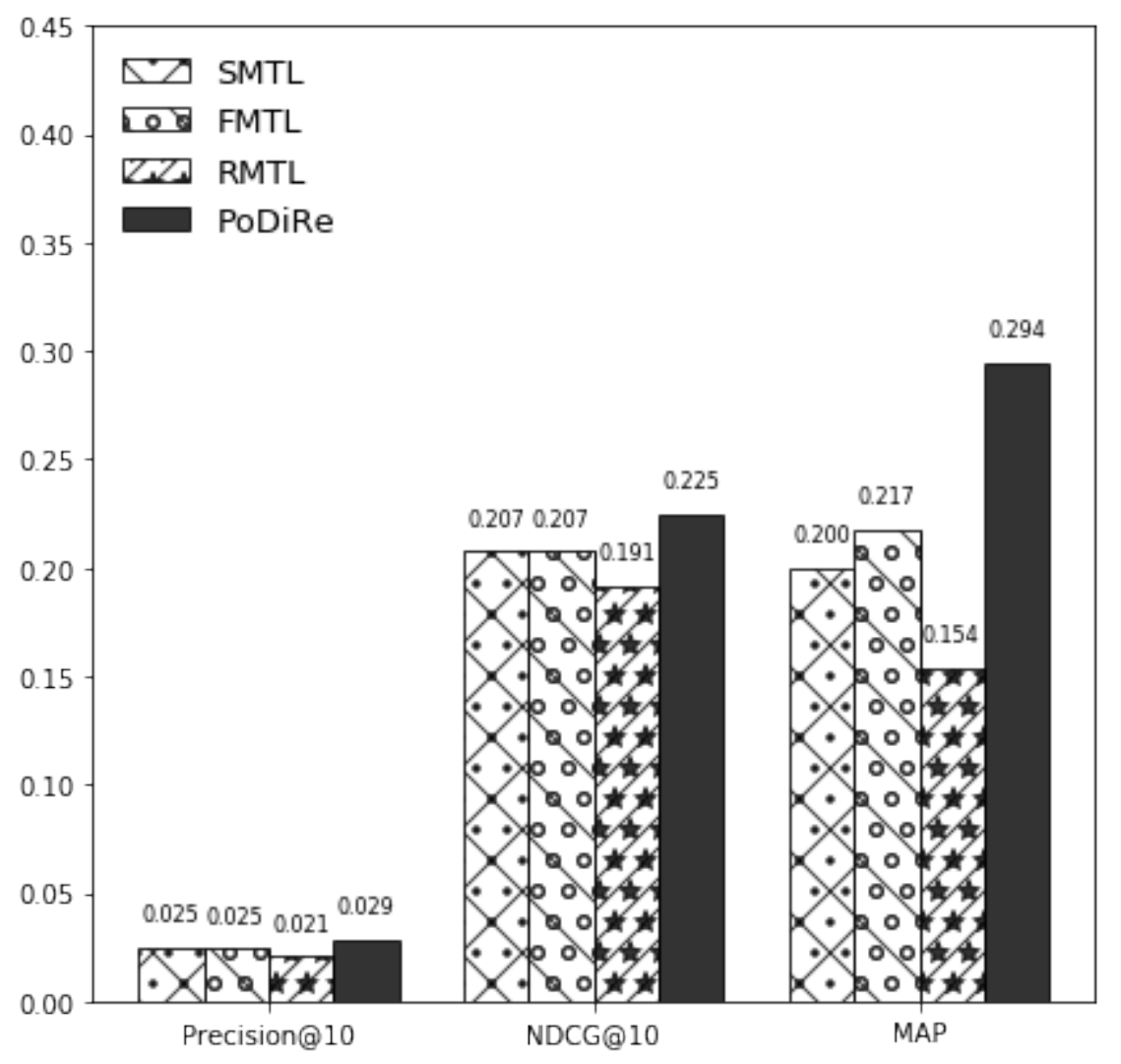}
\label{fig:comparison with multi-task supervised learning K=10 - install}}
\hfill
\subfigure[Play]{
\includegraphics[width=0.31\textwidth]{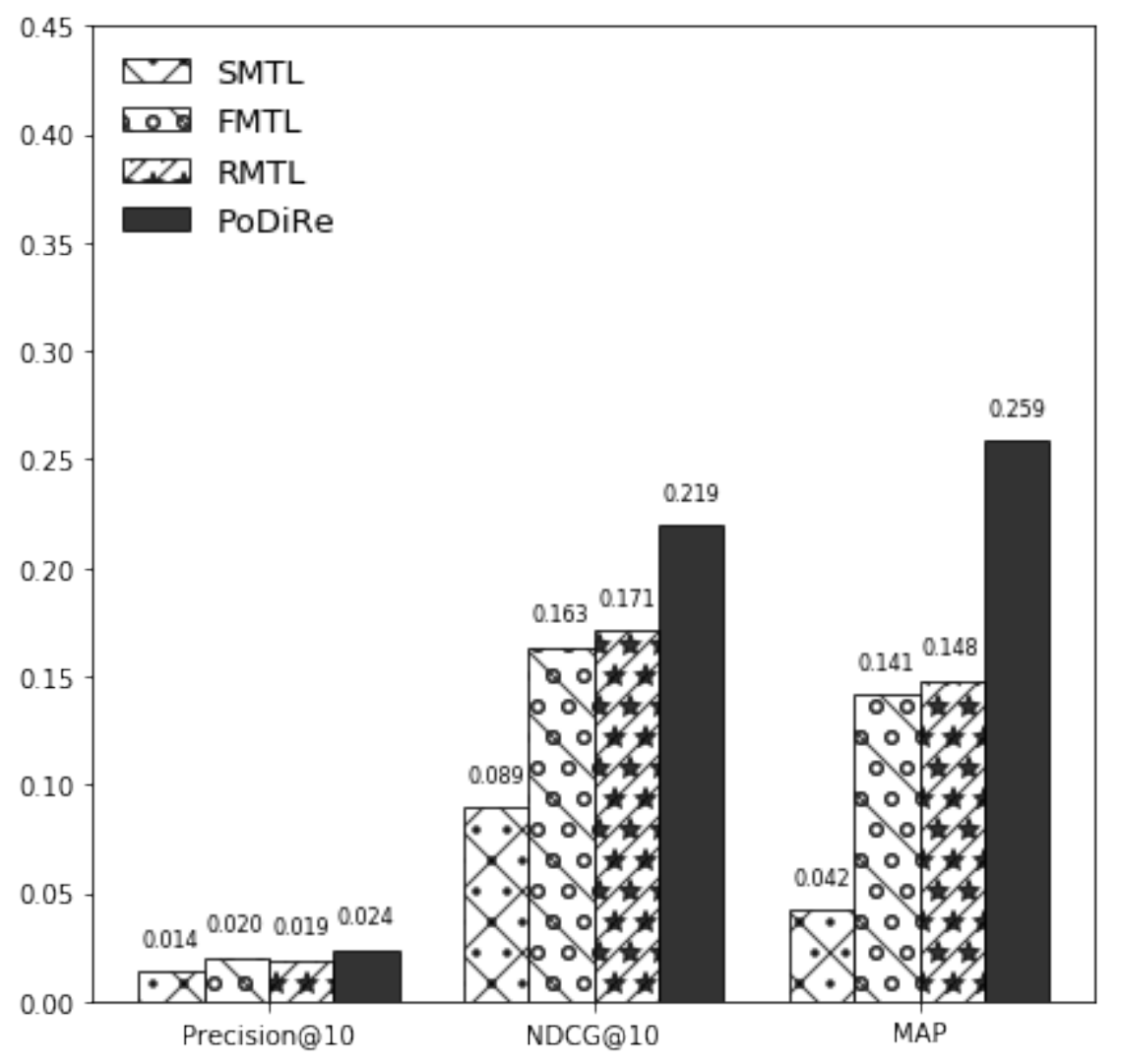}
\label{fig:comparison with multi-task supervised learning K=10 - play}}
\caption{Performance comparison between {\it PoDiRe} and major competitors based on multi-task supervised learning over multiple recommendation tasks with $K = 10$}
\label{fig:comparison with multi-task supervised learning K=10}
\end{figure*}
